\documentclass[sigconf]{acmart}

\usepackage{enumitem}
\usepackage{hyperref}
\usepackage{amsmath} 
\usepackage{multirow}
\usepackage{graphicx}
\usepackage{subfigure}
\usepackage{bm}
\usepackage{tabularx}
\usepackage{framed}
\usepackage{lipsum}
\usepackage{xcolor}
\usepackage{stfloats}

\usepackage{color}

\definecolor{sgadecikir}{rgb}{0.2,0.2,0.22}

\newcommand\blfootnote[1]{%
  \begingroup
  \renewcommand\thefootnote{}\footnote{#1}%
  \addtocounter{footnote}{-1}%
  \endgroup
}
\AtBeginDocument{%
  \providecommand\BibTeX{{%
    \normalfont B\kern-0.5em{\scshape i\kern-0.25em b}\kern-0.8em\TeX}}}

\acmConference[WWW '24]{the ACM Web Conference 2024}{May 13--17,
  2024}{Singapore}
\copyrightyear{2024}
\acmYear{2024}
\setcopyright{acmlicensed}\acmConference[WWW '24]{Proceedings of the ACM Web Conference 2024}{May 13--17, 2024}{Singapore, Singapore}
\acmBooktitle{Proceedings of the ACM Web Conference 2024 (WWW '24), May 13--17, 2024, Singapore, Singapore}
\acmDOI{10.1145/3589334.3645378}
\acmISBN{979-8-4007-0171-9/24/05}

\begin{document}



\title{UrbanCLIP: Learning Text-Enhanced Urban Region Profiling with Contrastive Language-Image Pretraining from the Web}


\author{Yibo Yan}
\affiliation{%
  \institution{Hong Kong University of Science and Technology (Guangzhou)}
  \city{}
  \country{}}
\email{yanyibo70@gmail.com}

\author{Haomin Wen}
\affiliation{%
  \institution{Beijing Jiaotong University}
  \city{}
  \country{}}
\email{wenhaomin@bjtu.edu.cn}

\author{Siru Zhong}
\affiliation{%
  \institution{Hong Kong University of Science and Technology (Guangzhou)}
  \city{}
  \country{}}
\email{zhongsiru204@gmail.com}

\author{Wei Chen}
\affiliation{%
  \institution{Hong Kong University of Science and Technology (Guangzhou)}
  \city{}
  \country{}}
\email{onedeanxxx@gmail.com}

\author{Haodong Chen}
\affiliation{%
 \institution{Northwestern Polytechnical University}
 \city{}
 \country{}}
\email{chd@mail.nwpu.edu.cn}

\author{Qingsong Wen}
\affiliation{%
  \institution{Alibaba Group, US}
  \city{}
  \state{}
  \country{}}
\email{qingsongedu@gmail.com}

\author{Roger Zimmermann}
\affiliation{%
  \institution{National University of Singapore}
  \country{}}
\email{rogerz@comp.nus.edu.sg}

\author{Yuxuan Liang$^{\dag}$}
\affiliation{%
  \institution{Hong Kong University of Science and Technology (Guangzhou)}
  \city{}
  \country{}}
\email{yuxliang@outlook.com}
\renewcommand{\shortauthors}{Yibo Yan et al.}


\begin{abstract}

Urban region profiling from web-sourced data is of utmost importance for urban computing. We are witnessing a blossom of LLMs for various fields, especially in multi-modal data research such as vision-language learning, where text modality serves as a supplement for images. As textual modality has rarely been introduced into modality combinations in urban region profiling, we aim to answer two fundamental questions: i) \emph{Can text modality enhance urban region profiling? ii) and if so, in what ways and which aspects?} To answer the questions, we leverage the power of Large Language Models (LLMs) and introduce the first-ever LLM-enhanced framework that integrates the knowledge of text modality into urban imagery, named LLM-enhanced \underline{Urban} Region Profiling with \underline{C}ontrastive \underline{L}anguage-\underline{I}mage \underline{P}retraining (\textbf{UrbanCLIP}). Specifically, it first generates a detailed textual description for each satellite image by Image-to-Text LLMs. Then, the model is trained on image-text pairs, seamlessly unifying language supervision for urban visual representation learning, jointly with contrastive loss and language modeling loss. Results on urban indicator prediction in four major metropolises show its superior performance, with an average improvement of 6.1\% on $R^2$ compared to the state-of-the-art methods. Our code and dataset are available at \url{https://github.com/StupidBuluchacha/UrbanCLIP}.
\blfootnote{$^{\dag}$Y. Liang is the corresponding author. Email: yuxliang@outlook.com} 


\end{abstract}

\maketitle

\section{INTRODUCTION}

The rapid pace of urbanization has led to more than half of the global population, totaling 4.4 billion inhabitants \cite{owidurbanization, worldbank2022}. \emph{Urban region profiling}, a pervasive and enduring theme within the domains of web mining and knowledge discovery, is the process of representing and summarizing key features and attributes of urban areas in a lower-dimensional space. By harnessing diverse web-sourced data, such as satellite \cite{cong2022satmae,huang2021m3g,han2020learning,han2020lightweight,wang2020urban2vec,yeh2020using} and street-view imagery \cite{li2022predicting,liu2023knowledge,wang2020urban2vec}, this process delivers a comprehensive understanding of urban spaces, spanning the realms of social, economic, and environmental aspects. In this way, urban region profiling empowers decision-makers with critical insights and related web systems into urban planning, sustainable development, and policy formulation.

\begin{figure}[!t]
  \centering
      \includegraphics[width=\linewidth]{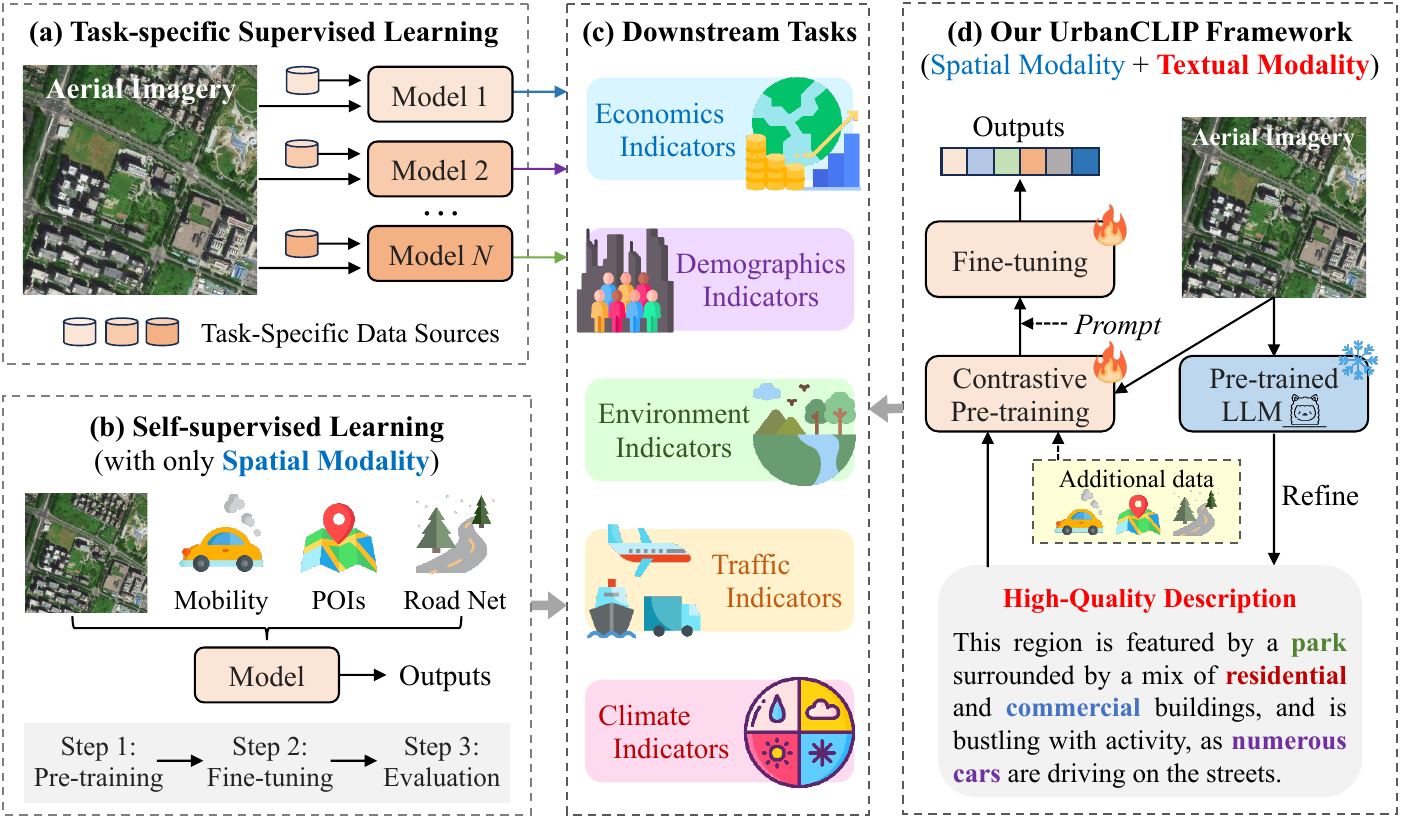}
  \vspace{-1.5em}
  \caption{Frameworks for urban region profiling. We present the first attempt to leverage the power of LLMs for this task.}
  \label{fig: intro}
  \vspace{-2em}
\end{figure}

Scholars and policymakers traditionally rely on manual surveys to gather urban statistics. However, such methods inherently face limitations in balancing high spatial resolution and real-time updates due to their prohibitive costs \cite{un2021sustainable,measureDHS2013,yeh2020using}. In contrast, data originating from web platforms boasts consistent updates and easy accessibility, especially high-resolution urban surfaces extracted from Baidu Map or Google Map \cite{li2022predicting,liu2023knowledge,xi2022beyond}, serving as the foundation for machine learning models to achieve a cost-friendly, high-quality, and timely understanding of urban indicators \cite{burke1, li2022predicting, 10.1145/3184558.3186581}. Upon revisiting the existing literature, we classify \emph{web-based} urban region profiling into two categories, as shown in Figure \ref{fig: intro}:

\par a) \textbf{Task-specific supervised learning} acquires urban region representations through supervised training using data sources (e.g., satellite imagery) specific to particular tasks, including poverty levels \cite{yeh2020using,jean2016combining,han2020learning,perez2017poverty,ayush2020generating,ayush2021efficient}, crop yields \cite{you2017deep,wang2018deep,russwurm2020self,martinez2021fully,m2019semantic,yeh2021sustainbench}, population, land cover \cite{hong2020graph,uzkent2019learning} and commercial activeness \cite{he2018perceiving,liu2023knowledge}. However, the task-specific nature of supervised learning, which requires considerable labeled data, may impede the model's \emph{generalizability}, potentially compromising its overall robustness and efficacy.

\par b) \textbf{Self-supervised learning}, extending beyond satellite imagery, integrates diverse auxiliary spatial modalities to generate comprehensive feature representations. These representations boast wide applicability, readily generalizing across numerous urban indicator tasks, as delineated in Figure \ref{fig: intro}(c). Typically, 
\cite{xi2022beyond,liu2023knowledge,bai2023geographic,jenkins2019unsupervised} integrate the information of Point-of-Interests (POIs) to capture human-inhabited areas and associated activities. Similarly, a series of studies consider aspects like mobility \cite{jenkins2019unsupervised,chen2022MainTUL} and human trajectory data to enhance urban region profiling \cite{li2022predicting,yang2022duare}. Nevertheless, these approaches often lack sufficient \emph{explanatory significance}, such as explaining in language that can easily be understood by humans.

During the past year, there has been a notable upsurge in the use of LLMs across various fields~\cite{chen2021evaluating,thoppilan2022lamda,ahn2022can,jintime2023}. The success is attributed to their remarkable proficiency in language understanding and the extensive knowledge they acquire during pre-training. Particularly, LLMs play a pivotal role in advancing multimodal learning, where textual data complements other modalities. As an example, the integration of rich textual information has proven beneficial in tasks like image captioning \cite{radford2021learning, tewel2022zerocap, zeng2023conzic} and video question-answering \cite{yang2022zero,su2023language,pan2023retrieving}. However, the incorporation of the textual modality in conjunction with urban imagery is a relatively unexplored area. Inspired by the significant achievements of LLMs in general fields, we embark on the exploration of two fundamental questions -- \textbf{Q1:} \emph{Can the inclusion of textual data serve as a powerful complement to satellite imagery for more effective urban region profiling?} \textbf{Q2:} \emph{and if so, in what ways and with regard to which specific aspects?}

To answer the aforementioned questions, we integrate the textual modality into urban imagery profiling for the first time, leading to a novel framework, named LLM-enhanced \underline{Urban} Region Profiling with \underline{C}ontrastive \underline{L}anguage-\underline{I}mage \underline{P}retraining, termed as UrbanCLIP. At first, we generate a detailed description by a well-trained LLM (LLaMA-Adapter V2 \cite{gao2023llama}) for each satellite image. Then, the high-quality image-text pairs are fed into UrbanCLIP with an encoder-decoder architecture.  It encodes satellite images to latent representation by a visual encoder (vision transformer \cite{dosovitskiy2020image}) and decodes texts with a causal masking transformer decoder. We further design a decoupled decoder mechanism, where unimodal textual representations from the first half of decoder layers would cascade the rest of decoder layers, cross-attending to the image encoder for multimodal representations.
Moreover, a contrastive loss is applied between unimodal image and text embeddings, while language modeling loss on the multimodal decoder outputs is utilized for natural language profiling of urban regions with detailed granularity. The text-incorporated visual representations can support the prediction of various urban indicators from different urban regions. Overall, the main contributions of our work are summarized as:

\begin{itemize}[leftmargin=*]

    \item Powered by LLM, UrbanCLIP is the first-ever framework that integrates the knowledge of text modality into urban region profiling. We show that such comprehensive textual data generated by pre-trained image-to-text LLM is a critical supplement to urban region representations.

    \item UrbanCLIP infuses textual knowledge into visual representations through deep modality interaction jointly with contrastive loss and language modeling loss, via a contrastive learning-based encoder-decoder architecture, which subsumes model capacities from both contrastive models and generative models. 

    \item Extensive experiments on four cities and three urban indicators demonstrate the effectiveness of UranCLIP.  Further analyses are conducted to show the transferability and interpretability of the proposed model. We also develop a web-based system to offer insights about urban computing, with an interactive experience. 

\end{itemize}

\vspace{-0.5em}
\section{PRELIMINARIES}
\subsection{Formulation}

\noindent \textbf{Definition 1 (Urban Region)} 
We follow prior studies \cite{xi2022beyond,liu2023knowledge} to partition an area of interest (\emph{e.g.}, a city) evenly into $L$ urban regions. 

\noindent \textbf{Definition 2 (Satellite Image)}
Based on the real-time monitoring of the Earth's surface by satellites, satellite imagery offers a comprehensive view of the structural characteristics of a given region. Each input satellite image w.r.t. an urban region $g$ can be denoted as $I_g \in \mathbb{R}^{H \times W \times 3}$, where $H$ and $W$ are length and width.

\noindent \textbf{Definition 3 (Location Description)} 
The description text $T_{g}$ for an urban region $g$ contains several individual sentences. Such text can be generated manually or using image captioning tools. E.g., by leveraging the well-trained LLM's profound understanding of general-purpose knowledge \cite{zhao2023survey,hu2023survey,li2022pretrained,wei2022emergent}, we can generate the summary text of a given region, especially including its spatial context (e.g., POIs) that significantly reflects its land function \cite{gao2023llama}. 

\noindent \textbf{Definition 4 (Urban Indicator)} 
Urban indicators gauge the urban region's standing on the socioeconomic spectrum and the environmental perspective. The $K$ indicators on a set of $L$ urban regions are denoted as $\textbf{Y} \in \mathbb{R}^{L \times K}$. In this paper, we use \emph{population} (\#citizens), \emph{GDP} (million Chinese Yuan), and \emph{carbon emission} (ton) as social, economic, and environmental ground-truth indicators, respectively.

\noindent \textbf{LLM-Enhanced Urban Region Profiling}.
 Given the above settings, we aim to learn a function $\mathcal{F}$ to map the \emph{satellite imagery}, its \emph{text description}, and other available data (e.g., POIs, road networks) to a vector $\boldsymbol{e}_g={\mathcal{F}}\left(I_g, T_g\right)$. The representations can be further used to infer urban indicators $\textbf{Y} \in \mathbb{R}^{L \times K}$ for an arbitrary set of regions.

\vspace{-0.5em}
\subsection{Related Work}

\subsubsection{Urban Region Profiling}

\par Learning urban region profiling from the web data has been a long-standing research topic in web mining. Current efforts can be broadly classified into two types: 

\par - \emph{Task-specific supervised learning.} This line of research learns prediction models from task-specific data sources. For example, using light intensity as supervision data, \citet{yeh2020using} employ a pre-trained CNN model to predict asset levels in Africa. Similar methodologies have been applied in forecasting economic indicators in studies like \cite{park2022learning, he2018perceiving, huang2021m3g}. Additionally, certain investigations estimate house prices by leveraging learned visual features from both satellite and street-view images, as seen in \cite{law2019take}.

\par - \emph{Self-supervised learning (SSL) with spatial modality.} This research strand mostly focuses on combining urban imagery and spatial modality for urban region profiling. They typically resort to Tobler's First Law of Geography \cite{miller2004tobler}, known as ``Everything is related to everything else, but near things are more related than distant things'', to distill urban imagery representations, via various similarity metrics \cite{jean2019tile2vec,wang2020urban2vec} or loss forms \cite{bjorck2021accelerating,kang2020deep,xi2022beyond}. Some studies, such as \cite{xi2022beyond}, incorporate POI data in a contrastive-learning framework, aiming to ensure that satellite images associated with similar POI distributions exhibit a closer relationship in visual latent space. Furthermore, \cite{liu2023knowledge} introduces an urban knowledge graph and infuses such semantic embedding into visual representation learning of satellite imagery via contrastive learning. In general, SSL outperforms task-specific supervised learning in terms of generalization. 

Compared with SSL with spatial modality, UrbanCLIP introduces the textual modality as complementary information for urban region profiling with the first shot, leading to a more comprehensive, generalizable, and interpretable urban region representation.

\vspace{-0.3em}
\subsubsection{Large Language Model}
LLM are renowned for their ability to attain comprehensive language understanding the generation, which stems from their training on massive datasets and billions of parameters. Inspired by their impressive performance, there is a rising trend of incorporating LLMs in various fields, such as ChatBot \cite{thoppilan2022lamda,anil2023palm,shuster2022blenderbot}, coding \cite{gunasekar2023textbooks, li2022competition, chen2021evaluating}, and even time series forecasting \cite{jin2023timellm,zhou2023one}. However, the potential of LLMs remains largely untapped in the field of urban computing, including our region profiling task. More related work can be found in Appendix \ref{app:related}.


\begin{figure}[!b]
  \centering
  \vspace{-1em}
  \includegraphics[width=0.48\textwidth]{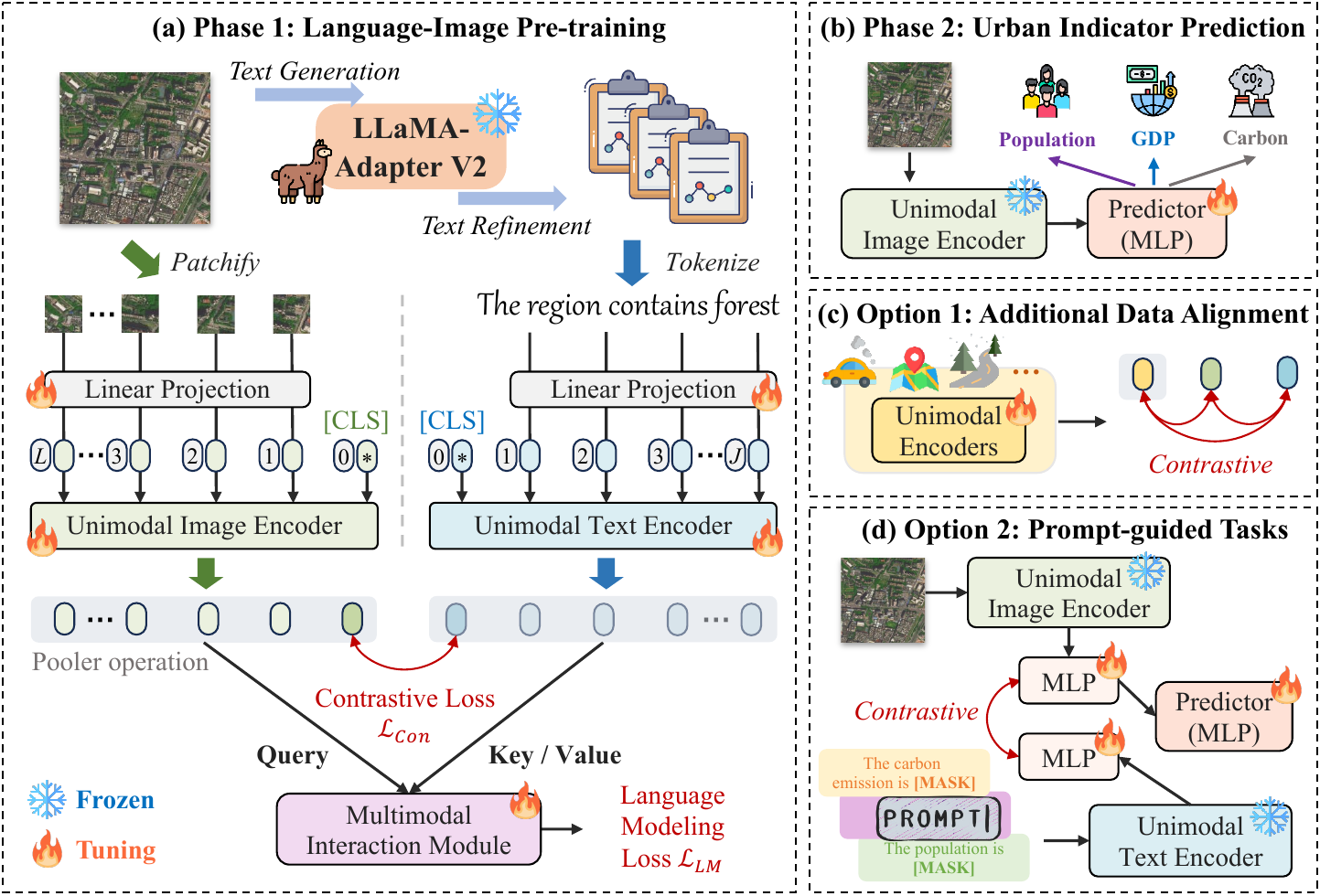}
  \vspace{-2em}
  \caption{Overall framework of the proposed UrbanCLIP.}
  \label{fig: framework}
\end{figure}

\vspace{-0.5em}
\section{METHODOLOGY}

As illustrated in Figure \ref{fig: framework}, the overall framework of UrbanCLIP is composed of two key phases with two optional settings. 
\begin{itemize}[leftmargin=*]
    \vspace{-0.2em}
    \item \textbf{Phase 1}: We first generate a detailed location description via LLaMA-Adapter V2 (an image-to-text foundation model) for the satellite imagery crawled from Baidu Map, thus forming a set of high-quality image-text pairs. The image and text are then fed into two unimodal encoders separately. Lastly, a multimodal interaction module is designed to align the representation of the two modalities in the latent space, with an elaborately designed cross-attention mechanism and contrastive learning objective.

    \item  \textbf{Phase 2}: In the urban indicator prediction phase, we utilize a frozen unimodal image encoder for downstream tasks, by simply fine-tuning outermost multi-layer perceptrons (MLPs) with a few trainable parameters. Furthermore, we offer two optional choices, which are a flexible infusion of other spatial modalities and prompt-guided urban indicator prediction.
\end{itemize}

\subsection{Text Generation and Refinement}

\begin{figure}[!h]
  \centering
  \vspace{-1em}
  \includegraphics[width=1 \linewidth]{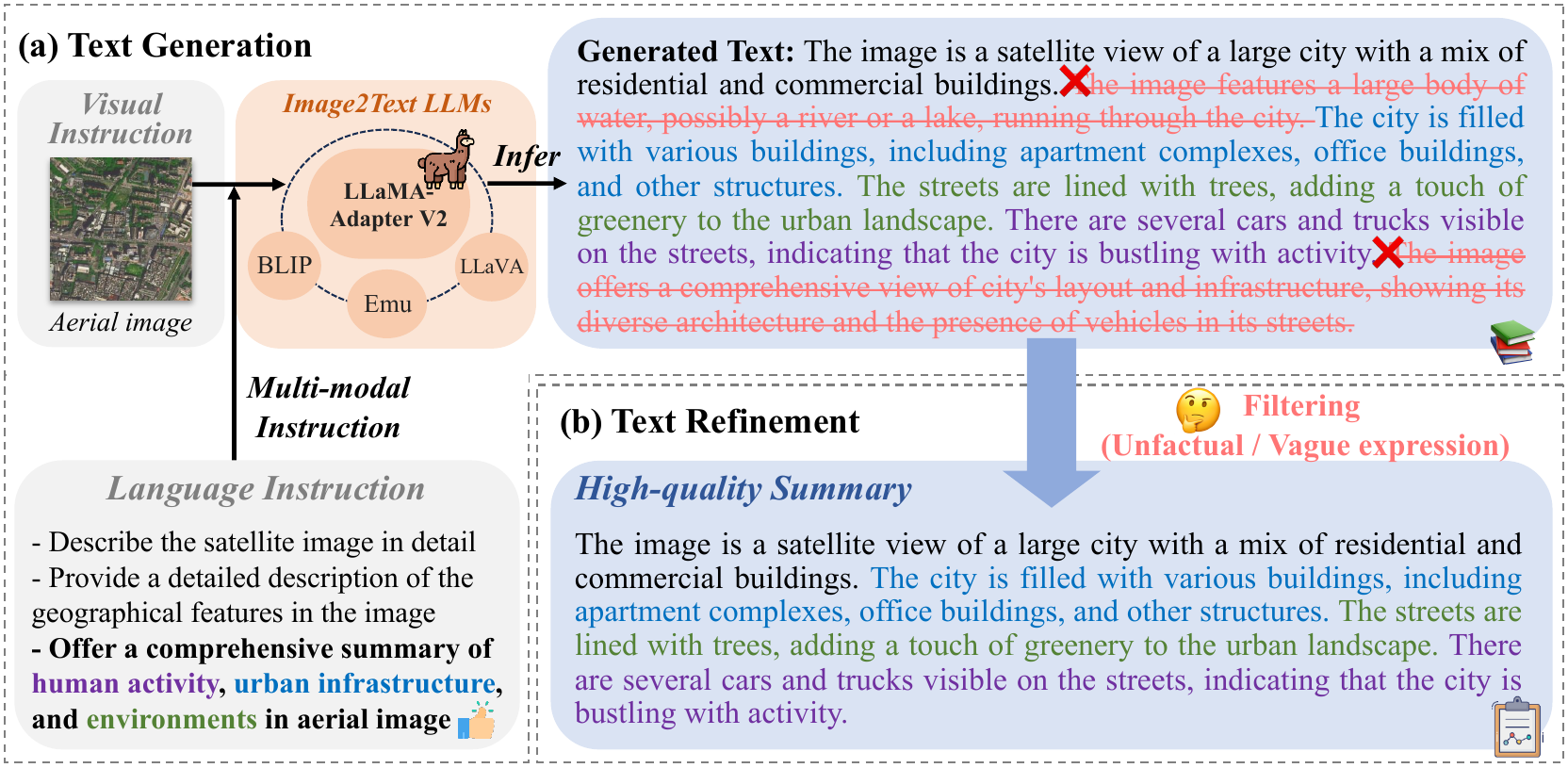}
  \vspace{-1.5em}
  \caption{Text generation and refinement.}
  \label{fig:text_gen}
  \vspace{-1em}
\end{figure}

\noindent \textbf{Text Generation.} For each satellite image, we adopt LLaMA-Adapter V2, an image-to-text foundation model, to generate a detailed location description as illustrated in Figure~\ref{fig:text_gen}(a). It takes a satellite image and an elaborately designed instruction as input and outputs a detailed text that describes the spatial information of the image. Through empirical experiments based on different language instructions, \emph{we find that a more detailed prompt, especially including a specific focus such as urban infrastructure, can trigger a more powerful capability of LLM to generate a high-quality summary}. We summarize other image-to-text counterparts in Appendix \ref{app: image2text}.

\noindent \textbf{Text Refinement.} As shown in the example of Figure \ref{fig:text_gen}(b), the generated description contains unfactual or vague information, and a thorough refinement, particularly the rule-based removal or rewriting, is conducted. As a result, a concise and high-quality summary retains the essential details about the satellite image, including its infrastructure, greenery, activity, etc. More details of the text refinement procedure are summarized in Appendix \ref{app: refinement}.

\subsection{Single-modality Representation Learning}
    \label{sec:single}
\textbf{Visual Representation Learning}. For an urban region $g$ with its satellite imagery $I_g$, we first split it into a sequence of patches $I_{p}$ (the default patch size is 16$\times$16), which are then linearly embedded into a dense vector: $ \boldsymbol{e}_{p}^{I} = \textbf{W}_{p} {I}_{p}^{\top} + {b}_{p}$, where $\textbf{W}_{p}$ and ${b}_{p}$ are learnable parameters. The learnable positional embeddings $\textbf{E}$ are further added to provide information about the relative position of each patch: $ \boldsymbol{e}_{E}^{I}=\boldsymbol{e}_{p}^{I}+\textbf{E}$. Then, $\boldsymbol{e}_{PE}^{I}$ is sent to the layers of the self-attention module to integrate the sequence information: $\left(\mathbf{Q}^{I}, \mathbf{K}^{I}, \mathbf{V}^{I} \right)^{\top} =\boldsymbol{e}_{PE}^{I} \left(\mathbf{W}_Q, \mathbf{W}_K, \mathbf{W}_{V}\right)^{\top}$,
where $\textbf{W}_Q, \textbf{W}_K$, and $\textbf{W}_V \in \mathbb{R}^{d \times d}$ are learnable matrices. The single-head and multi-head self attention (MSA) are defined as: 
\begin{equation}\label{eqn:MSA2}
\begin{split}
    \boldsymbol{e}^I_{(i)}=\operatorname{Softmax}\left(\textbf{Q}^{I} \textbf{K}^{{I}^{\top}} / \sqrt{d}\right) \textbf{V}^{I},\\
    \boldsymbol{e}^I_{MSA}=\operatorname{Concat}(\boldsymbol{e}^I_{(1)},\boldsymbol{e}^I_{(2)},...,\boldsymbol{e}^I_{(\#head)})\textbf{W}_{O},
\end{split}
\end{equation}
where $\textbf{W}_{O}$ is a learnable weight matrix, and $\operatorname{Concat}(\cdot)$ denotes the concatenation function. After residual connection and layer normalization, the latent visual representation can be obtained as: 
\begin{equation}    \boldsymbol{e}^{I}=\operatorname{LayerNorm}\left(\boldsymbol{e}^{I}_{E}+\operatorname{MSA}\left(\boldsymbol{e}^{I}_{E}\right)\right).
\end{equation}

It is noteworthy that the input satellite image patch sequence incorporates a learnable image [CLS] token at first to obtain dynamic interaction representation between patches. 
Inspired by~\cite{lee2019set}, we have implemented task-specific temporary pooling (abbreviated as pooler) to customize visual representation for distinct pre-training tasks while sharing the previous backbone encoder. The pooler serves as a task-specific self-attention layer, which acts as a natural task adapter. Specifically, we employ the overall sequence as the query attention pooler for the fine-grained cross-modality interaction task, and the [CLS] token as the query attention pooler for the coarse-grained cross-modality alignment task.

\noindent \textbf{Textual Representation Learning}.
For an urban region $g$, a high-quality text summary $T_g$ is generated from LLMs through text generation and refinement. Similar to the prior visual representation, it is desirable to encode this summary into a latent textual representation. Normally, BERT-style~\cite{kenton2019bert} models with encoder-only architecture can be generalized to capture latent textual representations. However, such traditional bidirectional attention may encounter low-rank issues~\cite{dong2021attention}, potentially weakening the model's expressive capacity and yielding limited generative capabilities. Hence, we choose a decoder-only architecture for the text encoding module. The primary distinction is that the textual representation is acquired via causally masked multi-head self-attention:
\begin{equation}    \boldsymbol{e}^{T}=\operatorname{LayerNorm}\left(\boldsymbol{e}^{T}_{E}+\operatorname{M-MSA}\left(\boldsymbol{e}^T_{E}\right)\right),
\end{equation}
where $\operatorname{M-MSA}$ means masked multi-head self attention operation, and $\boldsymbol{e}^{T}_{E}$ is the token representation of added location information. We also add a learnable [CLS] token to obtain the global information.

\vspace{-0.5em}
\subsection{Cross-modality Representation Learning}\label{sec:cross}
\textbf{Modality Alignment Task}.
While more visual tokens can help multi-modal understanding tasks, visual embeddings of image [CLS] tokens as global representations are beneficial for visual recognition and alignment tasks~\cite{yu2022coca}. 
Specifically, for the underlying visual representation learning sequence, we obtain a new sequence representation through self-attention and pooling operation:
\begin{equation}
    \boldsymbol{e}^{I}_{pool}=\operatorname{Pooling}(\operatorname{Softmax}(\boldsymbol{e}^{I}_{q}\boldsymbol{E}^{{I}^{\top}}_{k}) \cdot \boldsymbol{E}^{I}_{v}),
\end{equation}
where $\boldsymbol{E}^{I}_{k}$ equals $\boldsymbol{E}^{I}_{v}$ represents the sequence of visual representations before transformation, and $\boldsymbol{e}^{I}_{q}$ represents global visual embedding of [CLS] image token to be queried. For $\operatorname{Pooling}$, we select the $\operatorname{Mean-Pooling}$ to capture the global information.

We propose an image-text contrastive loss $\mathcal{L}_\mathrm{Con}$, as both LLM-enhanced semantic representation (i.e., text embedding) and visual representation (i.e., satellite imagery representation) of the same urban region should be as close to one another as possible. It can maximize the agreement of representations learned across different modalities while capturing different relationships. Thus, the two unimodal encoders are jointly optimized by contrasting the image-text pairs against others in the sampled batch of $m$ samples:
\begin{equation}\label{eq:con_loss}
\small
\nonumber
\begin{aligned}
    &\mathcal{L}_{Con} = \mathcal{L}_{Con}^{\text {Image } \rightarrow \mathrm{Text}}+\mathcal{L}_{Con}^{\mathrm{Text} \rightarrow \text { Image }} \\
    &= -\log \frac{\exp \left(\operatorname{sim}\left({\boldsymbol{e}}^{I}_{pool}, {\boldsymbol{e}}^{T}\right)\right)}{\sum_{i=1}^m \exp \left(\operatorname{sim}\left({\boldsymbol{e}}^{I}_{pool}, {\boldsymbol{e}}^{T}_{i}\right)\right)} 
    -\log \frac{\exp \left(\operatorname{sim}\left({\boldsymbol{e}}^{T}, {\boldsymbol{e}}^{I}_{pool}\right)\right)}{\sum_{i=1}^m \exp \left(\operatorname{sim}\left({\boldsymbol{e}}^{T}, {\boldsymbol{e}}_{{pool}_i}^{I}\right)\right)},
\end{aligned}
\end{equation}
where $\operatorname{sim}(\cdot)$ is inner product; $\mathcal{L}_{con}^{\text {Image } \rightarrow \mathrm{Text}}$ and $\mathcal{L}_{con}^{\mathrm{Text} \rightarrow \text { Image }}$ are image-to-text and text-to-image contrastive losses, repsectively.

\noindent \textbf{Modality Interaction Task}.
Unlike the previous studies where cross-modality interaction is shallow (e.g., via dot product-based similarity) \cite{liu2023knowledge, li2022predicting}, UrbanCLIP emphasizes the deep inter-modal interaction learning through layers for a contextualized feature sequence. Motivated by \cite{kim2021vilt,yu2022coca}, Transformer-based decoder architecture is then leveraged to fuse unimodal visual and textual representations together as multimodal representations. Specifically, UrbanCLIP employs multimodal decoder layers to effectively learn joint image-text representations, by leveraging unimodal textual encoder outputs and employing cross-attention mechanisms towards image encoder outputs. 
The key difference between multimodal cross-attention and unimodal MSA is that cross-attention uses visual modality as a query and text modality as key and value. 

Besides, to generate a text description for comprehensive urban region profiling, we introduce a language modeling loss $\mathcal{L}_\mathrm{LM}$ that enables the model to predict the next tokenized texts autoregressively with fine granularity. Hence, the multimodal decoder can learn to maximize the conditional likelihood of the paired text \textit{T}: $\mathcal{L}_{\text {LM }}=-\sum_{l=1}^{L} \log P_\theta\left(T_l \mid T_{<l}, I\right)$.

\vspace{-0.5em}
\subsection{Urban Indicator Prediction}
\textbf{Pre-training Stage}.
UrbanCLIP enables both unimodal text and multimodal representations to be generated simultaneously. To achieve this, both image-text contrastive loss $\mathcal{L}_\mathrm{Con}$ and language modeling loss $\mathcal{L}_\mathrm{LM}$ are applied. We minimize the following objective function for model learning during pre-training stage:
\begin{equation}\label{eq:total_loss}
\mathcal{L}_\mathrm{Total}=\lambda_\mathrm{Con}\cdot\mathcal{L}_\mathrm{Con}+\lambda_\mathrm{LM}\cdot\mathcal{L}_\mathrm{LM},
\end{equation}
where $\lambda_\mathrm{Con}$ and $\lambda_\mathrm{LM}$ are loss weighting hyperparameters.

An additional notable benefit of the loss design lies in its training efficiency \cite{yu2022coca}. The decoupled autoregressive decoder enables high-efficiency computation of two training losses. Unidirectional language models, trained with causal masking on complete texts, allow the decoder to generate outputs for both contrastive and generative training objectives in a single forward propagation. In contrast, the bidirectional approach requires two passes \cite{li2021align}, which is more time-consuming. As for UrbanCLIP, most computation is shared between the two losses. See Appendix \ref{app: complexity} for more details.

\noindent \textbf{Prediction Stage}.
Through optimizing the loss function in Eq. \ref{eq:total_loss}, we can obtain the final text-enhanced visual representations $\boldsymbol{e}_g$ from the frozen image encoder. Given any satellite image $I_g$, we can use a simple MLP to predict urban indicators as $\textbf{Y}_g=\operatorname{MLP}\left(\boldsymbol{e}_g\right)$.

\clearpage
\subsection{Discussion}
\vspace{-0.2em}
\subsubsection{Additional Data Alignment and Integration}

In reality, other spatial modalities such as POIs \cite{xi2022beyond,liu2023pre,bai2023geographic,jenkins2019unsupervised} and trajectories \cite{li2022predicting,yang2022duare} may be available which can contribute to urban region profiling. Considering this, we improve the flexibility of UrbanCLIP from the following two aspects: i) better alignment among diverse modalities. As illustrated in Figure \ref{fig: framework}(c), multimodality contrastive learning shows great capability in learning joint representations,  by maximizing the agreement between semantically aligned examples (i.e., positive sample) across modalities while minimizing the agreement between non-aligned ones. For more modalities, an example of a positive sample could be the combination of a satellite image, a text description, the majority of POI categories as parks, and the road network of a given area. ii) better interaction with existing modalities. An intuitive way is adopting cross-attention mechanisms in UrbanCLIP. For instance, each modality engages in attention with every other modality, creating pairwise interactions. In summary, UrbanCLIP supports a flexible infusion with other modalities as a plug-and-play integration for better urban region profiling.

\vspace{-0.2em}
\subsubsection{Prompt-guided Downstream Tasks}
Prompting was proposed initially in the natural language processing domain, and it refers to the generation of task-relevant instructions to obtain the desired output from a pre-trained model \cite{liu2023pre,jiang2020can}. Hence, a simple, task-specific prompt can be designed manually as one option to boost the downstream prediction performance of UrbanCLIP. As illustrated in Figure \ref{fig: framework}(d), for the carbon emission prediction task, a simple prompt can be designed during fine-tuning as ``The carbon emission is \textit{[MASK]}'', guiding the model to concentrate on the environment-related spatial information for visual representation learning. Specifically, a trainable MLP module is employed to align and inject textual prompt information. Furthermore, motivated by recent prompt learning-based studies, language instructions could be learned by training discrete \cite{jiang2020can,gao2020making} or continuous \cite{lester2021power,li2021prefix} vectors, consequently steering the performance of downstream urban indicators prediction.

\vspace{-0.5em}
\section{EXPERIMENTS}
\vspace{-0.1em}

In this section, we conduct extensive experiments to investigate the following Research Questions (RQ):
\vspace{-0.2em}
\begin{itemize}[leftmargin=*]
    \item \textbf{RQ1}: Can UrbanCLIP outperform prior approaches and generalize well to various urban indicator tasks?
    \item \textbf{RQ2}: How does each component (e.g., textual modality, text refinement, training objectives) contribute to UrbanCLIP?
    \item \textbf{RQ3}: How is the transferability of UrbanCLIP across cities?
    \item \textbf{RQ4}: How do we envision the practicality of UrbanCLIP?
\end{itemize}

\vspace{-0.5em}
\subsection{Experimental Setup}
\vspace{-0.1em}
\subsubsection{Datasets}

The datasets used in this paper include satellite imagery, textual description, and three urban indicators for four representative cities in China: Beijing, Shanghai, Guangzhou, and Shenzhen. The satellite images obtained from Baidu Map API have a fixed size of $256\times256$ with a spatial resolution of around 13 meters per pixel, which leads to an area of approximately 1 $km^2$. The textual information for each satellite image is generated from LLaMA-Adapter V2 \cite{gao2023llama}, which has the most detailed and high-quality text generation compared with other up-to-date open-source Image-to-Text foundation models \cite{liu2023llava,li2022blip,li2023blip,sun2023generative,han2023imagebindllm,su2023pandagpt,awadalla2023openflamingo,li-etal-2022-mplug} via empirical experiment. 
There exists a one-to-many relationship between images and associated texts.  We filter out low-quality descriptions and then adopt a random selection to choose one high-quality summary text that matches each satellite image. The overall statistics of satellite imagery and textual description can be seen in Table \ref{tab: dataset_statistics}. 
As for urban indicator data, we collect population from WorldPop \cite{worldpop} as a social indicator, GDP from \cite{doi:10.1080/15481603.2016.1276705} as an economic indicator and carbon emission from ODIAC  \cite{oda2018open} as the environmental indicator. All urban indicators per grid cell are aligned with corresponding satellite imagery and converted into a logarithmic scale. In this paper, we randomly partition the dataset into 60\% for training, 20\% for validation, and 20\% for test.


\vspace{-0.3em}
\subsubsection{Baselines}
We compare UrbanCLIP with the following baselines in the field of urban imagery-based socioeconomic prediction:
\begin{itemize}[leftmargin=*]
    \item \textbf{Autoencoder} \cite{kramer1991nonlinear}. A neural network architecture that acquires representations from unlabeled satellite images as input, with the training objective of minimizing the reconstruction error.

    \item \textbf{PCA} \cite{tipping1999probabilistic}. Principal Component Analysis (PCA) is utilized to transform original satellite imagery into extended vectors and compute the first 10 principal components for each image.

    \item \textbf{ResNet}$\textbf{-}$\textbf{18} \cite{he2015deep}. It is a well-established deep learning model pre-trained on ImageNet. It directly transfers a model trained on natural imagery to satellite imagery.

    \item \textbf{Tile2Vec} \cite{jean2019tile2vec}. An unsupervised model that employs a triplet loss to learn the visual representations, with the goal of minimizing the similarity of proximate satellite image pairs, while maximizing the dissimilarity of distant pairs.
    

    \item \textbf{READ} \cite{han2020lightweight}. Representation Extraction over an Arbitrary District (READ) is a semi-supervised model that leverages limited labeled data and transfer learning methods on a partially-labeled dataset to extract robust and lightweight satellite image representations, utilizing a teacher-student network with pre-trained models.

    \item \textbf{PG\-SimCLR} \cite{xi2022beyond}. A satellite image representation method for its competitive performance in socioeconomic prediction, leveraging SimCLR \cite{chen2020simple} to encourage similar representations for grids with analogous facility distribution and geo-adjacency.
\end{itemize}

\begin{table}[!t]
\centering
\caption{Dataset statistics.}
\vspace{-3mm}
\label{tab: dataset_statistics}
\scalebox{0.8}{
\begin{tabular}{l|c|c|c|c} 
\hline
\multirow{2}{*}{\textbf{Dataset}} & \multicolumn{2}{c|}{\textbf{Coverage}} & \multirow{2}{*}{\begin{tabular}[c]{@{}c@{}}\textbf{\#Satellite}\\\textbf{Image}\end{tabular}} & \multirow{2}{*}{\begin{tabular}[c]{@{}c@{}}\textbf{\#Location}\\\textbf{Description}\end{tabular}}  \\ 
\cline{2-3}
& \begin{tabular}[c]{@{}c@{}}\textbf{Bottom-left}\end{tabular} & \begin{tabular}[c]{@{}c@{}}\textbf{Top-right}\end{tabular} && \\ 
\hline
Beijing & 39.75°N, 116.03°E & 40.15°N, 116.79°E & 4,592 & 20,642 \\
Shanghai & 30.98°N, 121.10°E & 31.51°N, 121.80°E & 5,244 & 23,455 \\
Guangzhou & 22.94°N, 113.10°E & 23.40°N, 113.68°E & 3,402 & 15,539 \\
Shenzhen & 22.45°N, 113.75°E & 22.84°N, 114.62°E & 4,324 & 18,113 \\
\hline
\end{tabular}}
\vspace{-5mm} 
\end{table}

\begin{table*}[htbp]
    \centering
    \caption{Urban indicators prediction results in four datasets. The best results are in bold, and the second-best results are underlined. The last row indicates the relative improvement in percentage.}\vspace{-3mm}
    \scalebox{0.78}{
    \begin{tabular}{c|cccc ccccc| ccccc cccc} \toprule 
\textbf{Dataset}&\multicolumn{9}{|c|}{\textbf{Beijing}}&\multicolumn{9}{c}{\textbf{Shanghai}}\\ \hline
\multirow{2}{*}{\textbf{Model}}&\multicolumn{3}{c}{\textbf{Carbon}}&\multicolumn{3}{c}{\textbf{Population}}&\multicolumn{3}{c}{\textbf{GDP}}&\multicolumn{3}{|c}{\textbf{Carbon}}&\multicolumn{3}{c}{\textbf{Population}}&\multicolumn{3}{c}{\textbf{GDP}}\\
&$R^2$&RMSE&MAE&$R^2$&RMSE&MAE&$R^2$&RMSE&MAE&$R^2$&RMSE&MAE&$R^2$&RMSE&MAE&$R^2$&RMSE&MAE\\\hline
Autoencoder&0.099 &0.936 &0.621 &0.094 &0.988 &0.712 &0.115 &1.603 &0.858 &0.119 &0.968 &0.617 &0.101 &0.967 &0.800 &0.077 &1.782 &0.900 \\
PCA&0.124 &0.921 &0.598 &0.109 &0.968 &0.700 &0.102 &1.696 &0.882 &0.123 &0.952 &0.588 &0.131 &0.958 &0.802 &0.103 &1.702 &0.890 \\
ResNet-18&0.393 &0.599 &0.411 &0.202 &0.858 &0.680 &0.203 &1.280 &0.758 &0.451 &0.512 &0.460 &0.233 &0.852 &0.692 &0.217 &1.297 &0.777 \\
Tile2Vec&0.599 &0.512 &0.468 &0.204 &0.813 &0.635 &0.182 &1.356 &0.792 &0.572 &0.462 &0.390 &0.249 &0.801 &0.620 &0.169 &1.380 &0.806 \\
READ&0.284 &0.678 &0.545 &0.301 &0.813 &0.632 &0.208 &1.281 &0.759 &0.399 &0.588 &0.527 &0.322 &0.801 &0.600 &0.229 &1.296 &0.773 \\
PG-SimCLR&\underline{0.613} &\underline{0.489} &\underline{0.360} &\underline{0.362} &\underline{0.799} &\underline{0.599} &\underline{0.317} &\underline{1.114} &\underline{0.688} &\underline{0.597} &\underline{0.442} &\underline{0.356} &\underline{0.410} &\underline{0.790} &\underline{0.584} &\underline{0.319} &\underline{1.181} &\underline{0.725} \\\hline 
UrbanCLIP&\textbf{0.662} &\textbf{0.327} &\textbf{0.302} &\textbf{0.407} &\textbf{0.788} &\textbf{0.589} &\textbf{0.319} &\textbf{1.102} &\textbf{0.684} &\textbf{0.652} &\textbf{0.331} &\textbf{0.300} &\textbf{0.429} &\textbf{0.778} &\textbf{0.578} &\textbf{0.320} &\textbf{1.119} &\textbf{0.702} \\ 
Improvement&8.11\%&33.22\%&16.00\%&12.35\%&1.39\%&1.69\%&0.73\%&1.04\%&0.62\%&9.28\%&25.12\%&15.73\%&4.59\%&1.54\%&1.06\%&0.38\%&5.28\%&3.06\%\\ \bottomrule 
\multicolumn{19}{c}{}\\ \hline 
\textbf{Dataset}&\multicolumn{9}{|c|}{\textbf{Guangzhou}}&\multicolumn{9}{c}{\textbf{Shenzhen}}\\\hline
\multirow{2}{*}{\textbf{Model}}&\multicolumn{3}{c}{\textbf{Carbon}}&\multicolumn{3}{c}{\textbf{Population}}&\multicolumn{3}{c}{\textbf{GDP}}&\multicolumn{3}{|c}{\textbf{Carbon}}&\multicolumn{3}{c}{\textbf{Population}}&\multicolumn{3}{c}{\textbf{GDP}}\\ 
&$R^2$&RMSE&MAE&$R^2$&RMSE&MAE&$R^2$&RMSE&MAE&$R^2$&RMSE&MAE&$R^2$&RMSE&MAE&$R^2$&RMSE&MAE\\\hline
Autoencoder&0.068 &0.992 &0.736 &0.163 &0.991 &0.833 &0.122 &1.753 &0.887 &0.099 &0.970 &0.704 &0.122 &0.989 &0.817 &0.093 &1.901 &0.899 \\
PCA&0.087 &0.989 &0.688 &0.179 &0.989 &0.812 &0.134 &1.693 &0.862 &0.133 &0.956 &0.677 &0.134 &0.977 &0.810 &0.087 &1.902 &0.899 \\
ResNet-18&0.388 &0.500 &0.513 &0.244 &0.883 &0.711 &0.215 &1.290 &0.791 &0.409 &0.556 &0.503 &0.250 &0.880 &0.701 &0.165 &1.398 &0.844 \\
Tile2Vec&0.482 &0.499 &0.501 &0.269 &0.855 &0.683 &0.173 &1.346 &0.799 &0.466 &0.501 &0.486 &0.289 &0.841 &0.649 &0.123 &1.500 &0.881 \\
READ&0.353 &0.589 &0.589 &0.301 &0.849 &0.633 &0.200 &1.289 &0.766 &0.378 &0.600 &0.551 &0.301 &0.811 &0.631 &0.186 &1.356 &0.823 \\
PG-SimCLR&\underline{0.503 }&\underline{0.401 }&\underline{0.401 }&\underline{0.370 }&\underline{0.823 }&\underline{0.603 }&\underline{0.309 }&\underline{1.109 }&\underline{0.702 }&\underline{0.523 }&\underline{0.412 }&\underline{0.417 }&\underline{0.386 }&\underline{0.791 }&\underline{0.610 }&\underline{0.290 }&\underline{1.172 }&\underline{0.741} \\\hline 
UrbanCLIP&\textbf{0.587 }&\textbf{0.390 }&\textbf{0.389 }&\textbf{0.388 }&\textbf{0.801 }&\textbf{0.602 }&\textbf{0.309 }&\textbf{1.109 }&\textbf{0.700 }&\textbf{0.597 }&\textbf{0.373 }&\textbf{0.387 }&\textbf{0.391 }&\textbf{0.791 }&\textbf{0.602 }&\textbf{0.293 }&\textbf{1.153 }&\textbf{0.734}\\ 
Improvement&16.77\%&2.65\%&3.02\%&4.89\%&2.70\%&0.10\%&0.10\%&0.04\%&0.37\%&14.12\%&9.58\%&7.27\%&1.48\%&0.04\%&1.39\%&0.86\%&1.65\%&0.96\% \\ \bottomrule 
    \end{tabular}}
    \label{tab:comparison}
    \vspace{-1em}
\end{table*}



\vspace{-0.3em}
\subsubsection{Metrics and Implementation}
To assess the prediction performance, we adopt three commonly used evaluation metrics: coefficient of determination ($R^2$), rooted mean squared error (RMSE), and mean absolute error (MAE) \cite{jean2016combining,xi2022beyond}. Higher $R^2$, and lower RMSE, MAE means better performance. As for the default implementation of UrbanCLIP, Vision Transformer (ViT) \cite{dosovitskiy2020image} and the first half of transformer decoder are applied to convert the satellite image and location description into their unimodal representations, respectively; and the rest of transformer decoder can be used for multimodal interaction to generate image-text representations. The parameter initialization follows the setting from \cite{ilharco_gabriel_2021_5143773,cherti2023reproducible}. Adam optimizer is chosen to minimize the training loss during parameter learning. A grid search on hyperparameters is conducted, where search ranges for learning rate and batch size are set as $\{$2$e^{-6}$, 2$e^{-5}$, 2$e^{-4}$, 2$e^{-3}$, 2$e^{-2}$$\}$ and $\{$4, 8, 16, 32, 64$\}$, respectively.

\vspace{-0.3em}
\subsection{RQ1: Performance Comparison}
We empirically evaluate the performance of different models on the four datasets. The experimental results are shown in Table \ref{tab:comparison}, from which we can obtain the following findings:

\textbf{i) UrbanCLIP consistently achieves the best performance across all the datasets}. It outperforms the best baseline, PG-SimCLR, by 7.06\%, 4.75\%, 7.25\% and 5.49\% in terms of $R^2$ for Beijing, Shanghai, Guangzhou and Shenzhen, respectively. Besides, the average performance gain of UrbanCLIP on RMSE and MAE are 7.02\% and 4.27\%, respectively. The results further prove the effectiveness of introducing the text modality into the urban region profiling.

\textit{ii) UrbanCLIP achieves promising results across all three urban indicators, with carbon emission being the best, followed by population, and GDP ranking last}. The average $R^2$ improvement percentages for carbon emission, population and GDP prediction are 12.07\%, 5.83\% and 0.52\%, respectively. A better performance in environmental indicators may come from the text-enhanced nature of UrbanCLIP, since the location summary containing key POIs such as parks can help indicate whether the corresponding region is environmentally friendly but cannot deduce the wealth class around that area. This insight inspires future work to leverage non-spatial information (such as economic-related time series) to enhance economic indicators' prediction performance.

\textit{iii) Existing satellite imagery-based prediction approaches still lack the capability to profile urban regions comprehensively}.  Taking the spatial correlations of regions into account, PG-SimCLR \cite{xi2022beyond} (the best baseline model) and Tile2Vec \cite{jean2019tile2vec} achieve competitive results among most prediction tasks compared to other baselines, which indicates that extra knowledge is beneficial for visual representation learning. Nevertheless, these methods may not capture crucial semantics in satellite imagery, such as significant POIs, where textual information can enhance understanding. 






\begin{figure}[b!]
  \centering
  \vspace{-1em}
  \includegraphics[width=\linewidth]{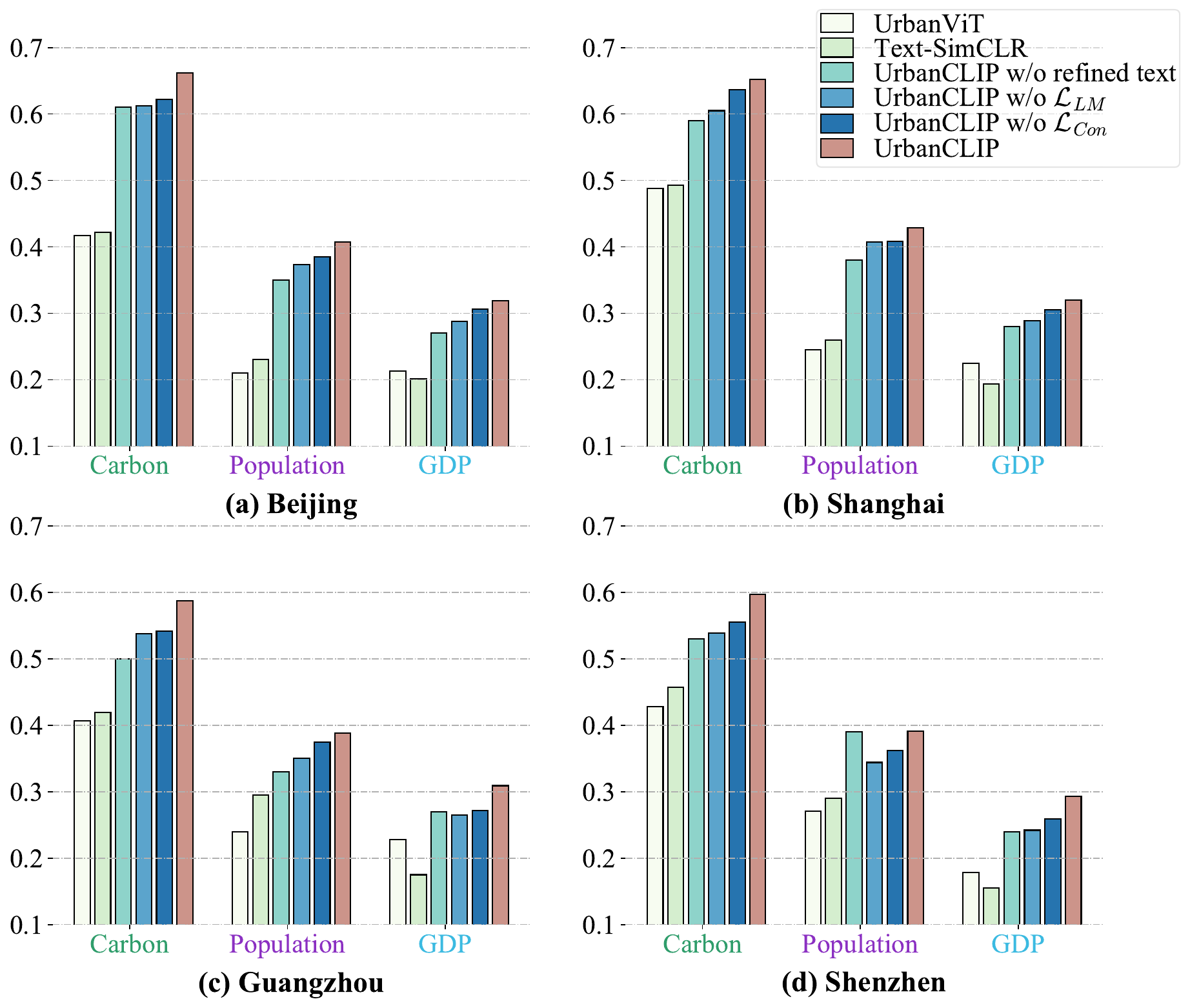}
  \vspace{-1em}
  \caption{Results of Ablation Study on $R^2$ Metric.}
  \label{fig: ablation}
\end{figure}

\begin{figure*}[!t]
  \centering
  \includegraphics[width=\textwidth]{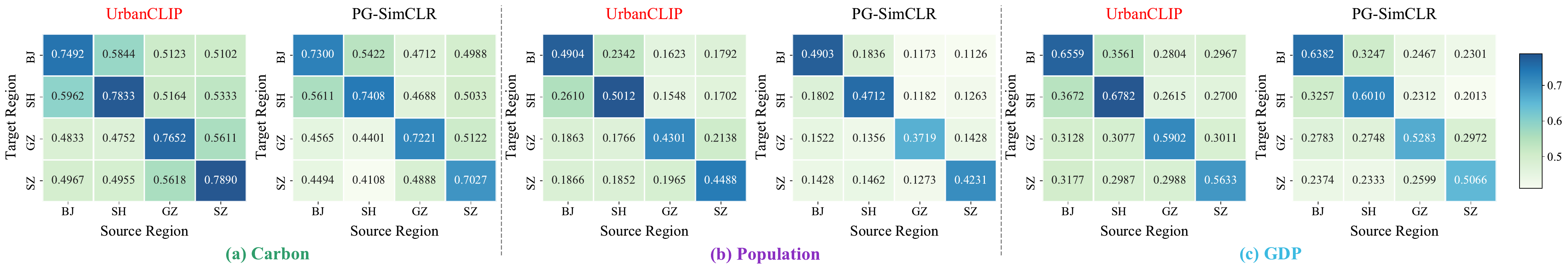}
  \vspace{-1.5em}
  \caption{$R^2$ heatmap for the transferability test between UrbanCLIP and PG-SimCLR, on 3 urban indicators across 4 datasets.}
  \label{fig: trans_all}
  \vspace{-1em}
\end{figure*}

\subsection{RQ2: Ablation Studies}
Next, we conduct ablation studies to investigate the effectiveness of different components in UrbanCLIP, including the generation and refinement of textual information,  cross-modality interaction, and training objectives. The results on $R^{2}$ are depicted in Figure \ref{fig: ablation}.

\subsubsection{Effectiveness of Textual Modality}
The core idea of UrbanCLIP is the introduction of textual modality for urban region profiling. Thus, it is natural to ask for the effectiveness of textual information. To this end, we compare UrbanCLIP with a standard ViT-based model \cite{dosovitskiy2020image}, termed as UrbanViT, which has the same setting as the unimodal visual encoder of UrbanCLIP. The extracted visual representations without textual enhancement would be used to predict three urban indicators. 

From Figure \ref{fig: ablation}, the absence of supplementary textual information (i.e., UrbanViT) results in significant performance deterioration, demonstrating the importance of textual modality for achieving a comprehensive visual representation. UrbanViT slightly outperforms ResNet-18 \cite{he2015deep}, which mainly comes from the powerful capability of ViT to capture global dependencies and contextual understanding in images \cite{dosovitskiy2020image, han2022survey}.

\vspace{-0.2em}
\subsubsection{Effectiveness of Refined Text}
Before feeding the text input into UrbanCLIP, we refine the generated satellite imagery summary for more robust model performance. To validate the effectiveness of this process, we report the performance of using raw generated summary (i.e., UrbanCLIP w/o refined text) for comparison.

Figure \ref{fig: ablation} clearly shows that UrbanCLIP consistently outperforms this variant across all cities and indicators, though the magnitude of the difference varies. Such result indicates that more relevant and noise-free textual information may align better with image features, leading to a more coherent and meaningful visual representation. To better understand the limitation of LLM, we summarize bad cases where LLM may not generate text effectively in Appendix \ref{app: bad case}.

\vspace{-0.2em}
\subsubsection{Effectiveness of Knowledge Infusion}
UrbanCLIP introduces contrastive learning-based cross-modality interaction coupled with image-text contrastive loss. To validate the efficacy of our approach in infusing textual knowledge, we introduce a direct image-image contrastive loss, denoted as Text-SimCLR, which is similar to PG-SimCLR \cite{xi2022beyond} (the best baseline). In particular, Text-SimCLR calculates textual embedding similarity for positive region pairs, and mandates that the associated satellite images of these pairs be proximate in the visual latent space.

Figure \ref{fig: ablation} shows the performance comparison between UrbanCLIP and Text-SimCLR over different datasets. The substantial performance gaps observed between these two models suggest that relying solely on the conventional image view-based contrastive loss fails to accomplish effective knowledge infusion. In particular, directly capturing the semantic knowledge inherent in location summaries as a similarity metric, yields a relatively weak self-supervision signal for visual representation learning. In contrast, our proposed cross-modality interaction mechanism, grounded in text-image contrastive learning, more effectively incorporates text-enhanced information within the multimodal representation space. In summary, the results highlight the efficacy of our proposed textual knowledge infusion, with potential applications extending to other research areas involving satellite imagery.

\vspace{-0.2em}
\subsubsection{Effectiveness of Loss Design}

We further investigate the effects of the two losses, i.e., image-text contrastive loss and language modeling loss. 
As depicted in Figure \ref{fig: ablation}, we assess the performance of UrbanCLIP in urban indicator prediction concerning contrastive-only and generative-only scenarios (denoted as UrbanCLIP w/o $\mathcal{L}_{LM}$ and w/o $\mathcal{L}_{Con}$, respectively) across four datasets. The findings reveal that, when compared to UrbanCLIP utilizing both losses, both single-loss variants exhibit relatively inferior $R^2$ performance. Furthermore, UrbanCLIP exclusively employing language modeling loss outperforms the counterpart with only contrastive loss. This observation implies that the generative objective contributes to refining text representations, thereby augmenting text comprehension for multimodal fusion with visual representations \cite{yu2022coca}. In essence, combining both losses fosters the acquisition of more semantically rich visual representations of satellite images.

\vspace{-0.3em}
\subsection{RQ3: Transferability Study}
We then focus on the transferability of UrbanCLIP, by investigating its performance on unseen regions (not included in training).
\subsubsection{Performance Across Cities}
We conduct experiments of UrbanCLIP and PG-SimCLR on metropolises in China with different geological and demographic characteristics: 1) Beijing, located in the northern part of China as the capital, is densely populated and characterized by a mix of traditional architecture and modern facilities; 2) Shanghai, situated on the eastern coast, serves as a global financial center known for its cosmopolitan atmosphere and iconic skyline; 3) Guangzhou, positioned in southern China, is a major trading and manufacturing center and has an intricate network of waterways; 4) Shenzhen has the almost same location distribution as Guangzhou, but it has transformed into a bustling metropolis characterized by technology parks and industrial zones. 

As shown in Figure \ref{fig: trans_all}, UrbanCLIP performs better than PG-SimCLR on 36 source-target pairs across three urban indicators. UrbanCLIP achieves an average $R^2$ of around 0.411, while that of PG-SimCLR is 0.365. Specifically, UrbanCLIP has higher $R^2$ values for respective urban indicators (carbon emission, population, and GDP) as 0.588, 0.384, and 0.261, but those of PG-SimCLR are only 0.543, 0.338, and 0.215. Such results indicate the stable transferability of our proposed UrbanCLIP in urban regions, although the chosen cities have the aforementioned differences in terms of geological and demographic characteristics.

The good transferability of our proposed UrbanCLIP may be attributed to our cross-modality mutual information maximization paradigm, through effective alignment and information preservation across visual representations and spatial semantics-enhanced textual representations. UrbanCLIP can better extract the inclusive functional semantics hidden behind satellite imagery, especially in urban scenarios involving spatial distribution shifts. Hence, although explicit differences exist among different cities, UrbanCLIP has the potential to address inaccuracies in the unseen satellite imagery of urban regions.

\subsubsection{Similarity Analysis Across Cities}

To better validate the transferability and explainability of UrbanCLIP across diverse urban regions, we compute the similarity between text-enhanced visual representations of satellite imagery. In particular, for a given satellite image from a source city, we compute the cosine similarity of visual representations among all others from different target cities. We assess whether there are commonalities in terms of urban indicators and description texts generated by UrbanCLIP. We further investigate the capability of generated text to guide associated image representations to focus on similar spatial information.

\begin{figure}[!t]
  \centering
  \includegraphics[width=1\linewidth]{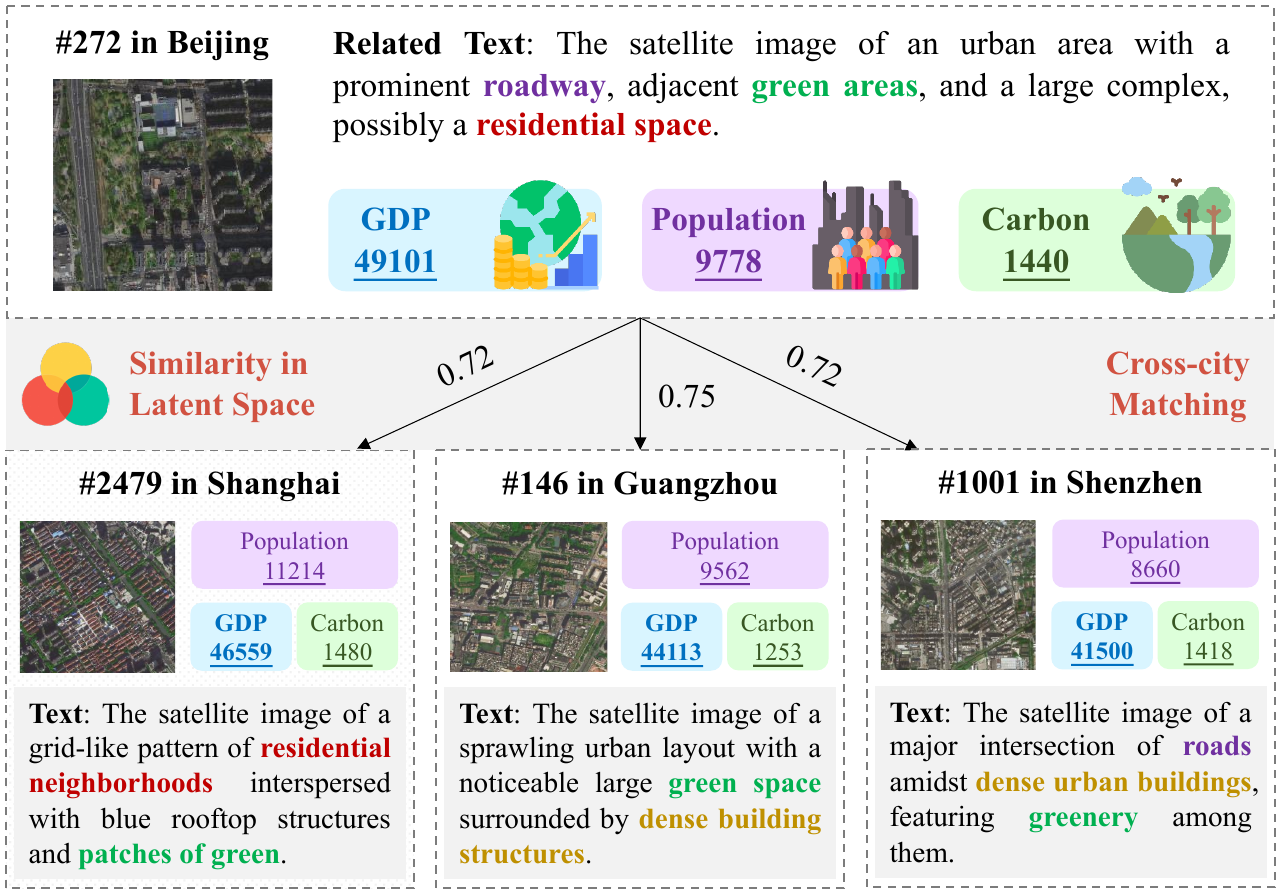}
  \caption{Case study of most similar satellite imagery matching between Beijing and other three cities through text-enhanced visual representations by UrbanCLIP.}
  \label{fig: similarity}
\end{figure}

As illustrated by Figure \ref{fig: similarity}, a randomly chosen satellite image in Beijing corresponds to three satellite images from other cities (Shanghai, Guangzhou, and Shenzhen) with the highest similarities (0.72, 0.75, and 0.72, respectively) in text-enhanced visual representations. In terms of urban indicators of regions corresponding to these satellite images, we can see that they are very close to each other. This phenomenon suggests that UrbanCLIP can capture similar spatial characteristics and distributions among comparable regions, thereby contributing to effective urban region profiling.



\subsection{RQ4: Practicality}
\begin{figure}[!t]
  \centering
  \includegraphics[width=0.48\textwidth]{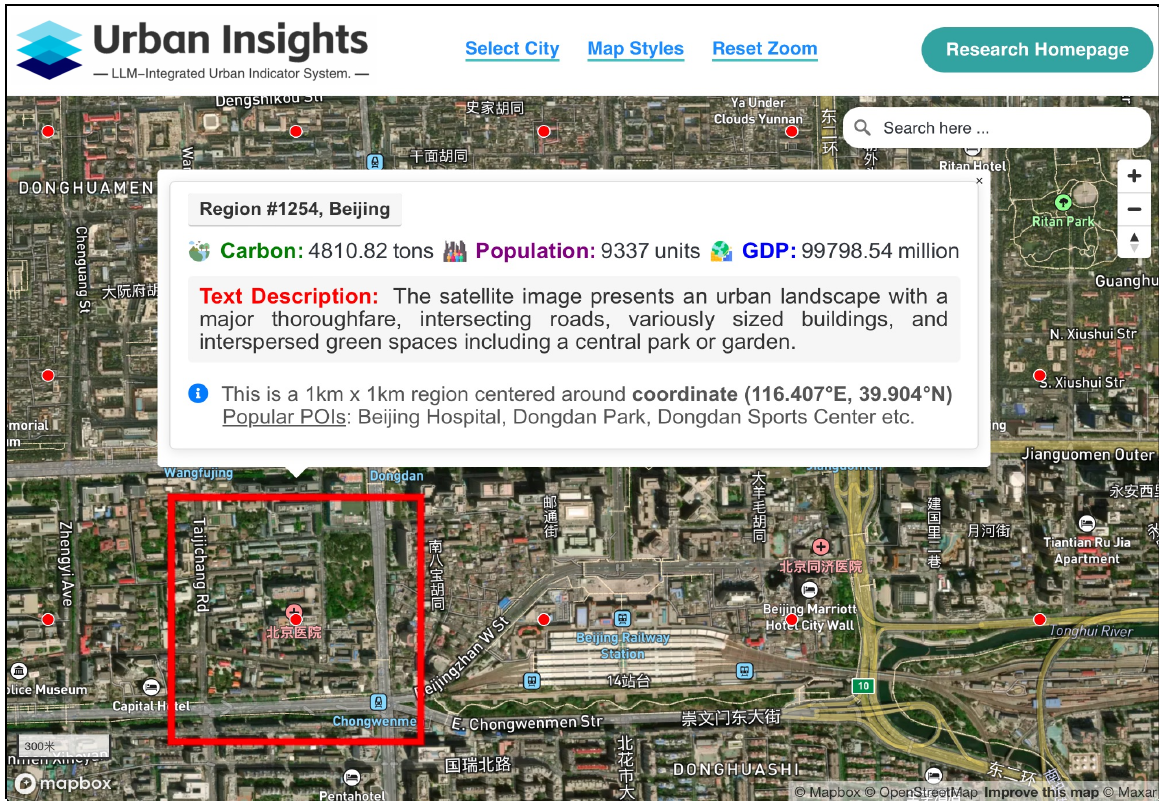}
  \caption{User interface of our Urban Insights System. It provides an interactive Mapbox-based platform \protect\cite{worldpop} for urban region query (e.g., showing image captions and POI queries) and profiling (i.e., calculating the urban indicators).}
  \label{fig: system}
\end{figure}

We finally envision and develop a novel web-based application called Urban Insights, which is an LLM-Integrated Urban Indicator System built on the Mapbox platform \cite{mapbox}. It displays urban landscapes in satellite projection, offering an interactive user experience. As shown in Figure \ref{fig: system}, users can easily navigate the map by zooming in and out, searching for special locations, and switching between different areas. Overlaid on this imagery are target grid areas, which will furnish users with detailed metrics, including carbon emissions, population, and GDP once clicked. Complementing the visual data, the system also features a descriptive image captioning module, which provides an easy-to-read text for understanding the spatial attributes of the selected grid. In addition, the system also supports popular POI query features within a region to better understand region functions. In summary, the Urban Insights System has great potential to provide users with a comprehensive view of varied urban landscapes and their prominent indicators, translating intricate urban data into a more accessible and intuitive visual representation. 


\section{CONCLUSION and FUTURE WORK}
\label{conclusion}

Profiling urban areas in terms of social, economic, and environmental metrics is critical for urban computing. 
This paper investigates whether and how the text modality benefits urban region profiling. To answer the question, we propose UrbanCLIP, the first-ever framework that integrates textual modality into urban imagery profiling. Powered by LLM, UrbanCLIP first generates a high-quality text description for an urban image. Subsequently, the text-image pairs are fed into the proposed model that seamlessly unifies natural language supervision for urban visual representation learning. Extensive experiments demonstrate the effectiveness of UrbanCLIP.

We aspire that this work motivates future research of urban region profiling on the following areas: 1) Investigating efficient and effective methods for integrating urban multimodal data and facilitating prompt-enhanced learning; 2) Exploring the automatic, high-quality text generation and refinement using more up-to-date LLMs; 3) Identifying more potentially beneficial downstream tasks, encouraging other researchers to explore diverse use cases.


\begin{acks}
This work is mainly supported by Guangzhou-HKUST (GZ) Joint Funding Program (No. 2024A03J0620). This work is also supported by the Advanced Research and Technology Innovation Centre (ARTIC), the National University of Singapore under Grant (project number: A-8000969-00-00).
\end{acks}

\bibliographystyle{ACM-Reference-Format}
\bibliography{urbanclip_reference}


\begin{thebibliography}{103}


\ifx \showCODEN    \undefined \def \showCODEN     #1{\unskip}     \fi
\ifx \showDOI      \undefined \def \showDOI       #1{#1}\fi
\ifx \showISBNx    \undefined \def \showISBNx     #1{\unskip}     \fi
\ifx \showISBNxiii \undefined \def \showISBNxiii  #1{\unskip}     \fi
\ifx \showISSN     \undefined \def \showISSN      #1{\unskip}     \fi
\ifx \showLCCN     \undefined \def \showLCCN      #1{\unskip}     \fi
\ifx \shownote     \undefined \def \shownote      #1{#1}          \fi
\ifx \showarticletitle \undefined \def \showarticletitle #1{#1}   \fi
\ifx \showURL      \undefined \def \showURL       {\relax}        \fi
\providecommand\bibfield[2]{#2}
\providecommand\bibinfo[2]{#2}
\providecommand\natexlab[1]{#1}
\providecommand\showeprint[2][]{arXiv:#2}

\bibitem[Ahn et~al\mbox{.}(2022)]%
        {ahn2022can}
\bibfield{author}{\bibinfo{person}{Michael Ahn}, \bibinfo{person}{Anthony
  Brohan}, \bibinfo{person}{Noah Brown}, \bibinfo{person}{Yevgen Chebotar},
  \bibinfo{person}{Omar Cortes}, \bibinfo{person}{Byron David},
  \bibinfo{person}{Chelsea Finn}, \bibinfo{person}{Chuyuan Fu},
  \bibinfo{person}{Keerthana Gopalakrishnan}, \bibinfo{person}{Karol Hausman},
  {et~al\mbox{.}}} \bibinfo{year}{2022}\natexlab{}.
\newblock \showarticletitle{Do as i can, not as i say: Grounding language in
  robotic affordances}.
\newblock \bibinfo{journal}{\emph{arXiv preprint arXiv:2204.01691}}
  (\bibinfo{year}{2022}).
\newblock


\bibitem[Anil et~al\mbox{.}(2023)]%
        {anil2023palm}
\bibfield{author}{\bibinfo{person}{Rohan Anil}, \bibinfo{person}{Andrew~M Dai},
  \bibinfo{person}{Orhan Firat}, \bibinfo{person}{Melvin Johnson},
  \bibinfo{person}{Dmitry Lepikhin}, \bibinfo{person}{Alexandre Passos},
  \bibinfo{person}{Siamak Shakeri}, \bibinfo{person}{Emanuel Taropa},
  \bibinfo{person}{Paige Bailey}, \bibinfo{person}{Zhifeng Chen},
  {et~al\mbox{.}}} \bibinfo{year}{2023}\natexlab{}.
\newblock \showarticletitle{Palm 2 technical report}.
\newblock \bibinfo{journal}{\emph{arXiv preprint arXiv:2305.10403}}
  (\bibinfo{year}{2023}).
\newblock


\bibitem[Awadalla et~al\mbox{.}(2023)]%
        {awadalla2023openflamingo}
\bibfield{author}{\bibinfo{person}{Anas Awadalla}, \bibinfo{person}{Irena Gao},
  \bibinfo{person}{Josh Gardner}, \bibinfo{person}{Jack Hessel},
  \bibinfo{person}{Yusuf Hanafy}, \bibinfo{person}{Wanrong Zhu},
  \bibinfo{person}{Kalyani Marathe}, \bibinfo{person}{Yonatan Bitton},
  \bibinfo{person}{Samir Gadre}, \bibinfo{person}{Shiori Sagawa},
  {et~al\mbox{.}}} \bibinfo{year}{2023}\natexlab{}.
\newblock \showarticletitle{Openflamingo: An open-source framework for training
  large autoregressive vision-language models}.
\newblock \bibinfo{journal}{\emph{arXiv preprint arXiv:2308.01390}}
  (\bibinfo{year}{2023}).
\newblock


\bibitem[Ayush et~al\mbox{.}(2020)]%
        {ayush2020generating}
\bibfield{author}{\bibinfo{person}{Kumar Ayush}, \bibinfo{person}{Burak
  Uzkent}, \bibinfo{person}{Marshall Burke}, \bibinfo{person}{David Lobell},
  {and} \bibinfo{person}{Stefano Ermon}.} \bibinfo{year}{2020}\natexlab{}.
\newblock \showarticletitle{Generating interpretable poverty maps using object
  detection in satellite images}.
\newblock \bibinfo{journal}{\emph{arXiv preprint arXiv:2002.01612}}
  (\bibinfo{year}{2020}).
\newblock


\bibitem[Ayush et~al\mbox{.}(2021)]%
        {ayush2021efficient}
\bibfield{author}{\bibinfo{person}{Kumar Ayush}, \bibinfo{person}{Burak
  Uzkent}, \bibinfo{person}{Kumar Tanmay}, \bibinfo{person}{Marshall Burke},
  \bibinfo{person}{David Lobell}, {and} \bibinfo{person}{Stefano Ermon}.}
  \bibinfo{year}{2021}\natexlab{}.
\newblock \showarticletitle{Efficient poverty mapping from high resolution
  remote sensing images}. In \bibinfo{booktitle}{\emph{Proceedings of the AAAI
  Conference on Artificial Intelligence}}, Vol.~\bibinfo{volume}{35}.
  \bibinfo{pages}{12--20}.
\newblock


\bibitem[Bai et~al\mbox{.}(2023)]%
        {bai2023geographic}
\bibfield{author}{\bibinfo{person}{Lubin Bai}, \bibinfo{person}{Weiming Huang},
  \bibinfo{person}{Xiuyuan Zhang}, \bibinfo{person}{Shihong Du},
  \bibinfo{person}{Gao Cong}, \bibinfo{person}{Haoyu Wang}, {and}
  \bibinfo{person}{Bo Liu}.} \bibinfo{year}{2023}\natexlab{}.
\newblock \showarticletitle{Geographic mapping with unsupervised multi-modal
  representation learning from VHR images and POIs}.
\newblock \bibinfo{journal}{\emph{ISPRS Journal of Photogrammetry and Remote
  Sensing}}  \bibinfo{volume}{201} (\bibinfo{year}{2023}),
  \bibinfo{pages}{193--208}.
\newblock


\bibitem[Bank(2022)]%
        {worldbank2022}
\bibfield{author}{\bibinfo{person}{The~World Bank}.}
  \bibinfo{year}{2022}\natexlab{}.
\newblock \bibinfo{booktitle}{\emph{Urban Development}}.
\newblock
\urldef\tempurl%
\url{https://www.worldbank.org/en/topic/urbandevelopment/overview}
\showURL{%
\tempurl}
\newblock
\shownote{Accessed: 2022-11-01}.


\bibitem[Bjorck et~al\mbox{.}(2021)]%
        {bjorck2021accelerating}
\bibfield{author}{\bibinfo{person}{Johan Bjorck}, \bibinfo{person}{Brendan~H
  Rappazzo}, \bibinfo{person}{Qinru Shi}, \bibinfo{person}{Carrie Brown-Lima},
  \bibinfo{person}{Jennifer Dean}, \bibinfo{person}{Angela Fuller}, {and}
  \bibinfo{person}{Carla Gomes}.} \bibinfo{year}{2021}\natexlab{}.
\newblock \showarticletitle{Accelerating ecological sciences from above:
  Spatial contrastive learning for remote sensing}. In
  \bibinfo{booktitle}{\emph{Proceedings of the AAAI Conference on Artificial
  Intelligence}}, Vol.~\bibinfo{volume}{35}. \bibinfo{pages}{14711--14720}.
\newblock


\bibitem[Burke et~al\mbox{.}(2021)]%
        {burke1}
\bibfield{author}{\bibinfo{person}{Marshall Burke}, \bibinfo{person}{Anne
  Driscoll}, \bibinfo{person}{David~B. Lobell}, {and} \bibinfo{person}{Stefano
  Ermon}.} \bibinfo{year}{2021}\natexlab{}.
\newblock \showarticletitle{Using satellite imagery to understand and promote
  sustainable development}.
\newblock \bibinfo{journal}{\emph{Science}} \bibinfo{volume}{371},
  \bibinfo{number}{6535} (\bibinfo{year}{2021}), \bibinfo{pages}{eabe8628}.
\newblock
\urldef\tempurl%
\url{https://doi.org/10.1126/science.abe8628}
\showDOI{\tempurl}


\bibitem[Chen et~al\mbox{.}(2021)]%
        {chen2021evaluating}
\bibfield{author}{\bibinfo{person}{Mark Chen}, \bibinfo{person}{Jerry Tworek},
  \bibinfo{person}{Heewoo Jun}, \bibinfo{person}{Qiming Yuan},
  \bibinfo{person}{Henrique Ponde de~Oliveira Pinto}, \bibinfo{person}{Jared
  Kaplan}, \bibinfo{person}{Harri Edwards}, \bibinfo{person}{Yuri Burda},
  \bibinfo{person}{Nicholas Joseph}, \bibinfo{person}{Greg Brockman},
  {et~al\mbox{.}}} \bibinfo{year}{2021}\natexlab{}.
\newblock \showarticletitle{Evaluating large language models trained on code}.
\newblock \bibinfo{journal}{\emph{arXiv preprint arXiv:2107.03374}}
  (\bibinfo{year}{2021}).
\newblock


\bibitem[Chen et~al\mbox{.}(2020)]%
        {chen2020simple}
\bibfield{author}{\bibinfo{person}{Ting Chen}, \bibinfo{person}{Simon
  Kornblith}, \bibinfo{person}{Mohammad Norouzi}, {and}
  \bibinfo{person}{Geoffrey Hinton}.} \bibinfo{year}{2020}\natexlab{}.
\newblock \showarticletitle{A simple framework for contrastive learning of
  visual representations}. In \bibinfo{booktitle}{\emph{International
  conference on machine learning}}. PMLR, \bibinfo{pages}{1597--1607}.
\newblock


\bibitem[Chen et~al\mbox{.}(2022)]%
        {chen2022MainTUL}
\bibfield{author}{\bibinfo{person}{Wei Chen}, \bibinfo{person}{Shuzhe Li},
  \bibinfo{person}{Chao Huang}, \bibinfo{person}{Yanwei Yu},
  \bibinfo{person}{Yongguo Jiang}, {and} \bibinfo{person}{Junyu Dong}.}
  \bibinfo{year}{2022}\natexlab{}.
\newblock \showarticletitle{Mutual Distillation Learning Network for
  Trajectory-User Linking}. In \bibinfo{booktitle}{\emph{IJCAI}}.
\newblock


\bibitem[Cherti et~al\mbox{.}(2023)]%
        {cherti2023reproducible}
\bibfield{author}{\bibinfo{person}{Mehdi Cherti}, \bibinfo{person}{Romain
  Beaumont}, \bibinfo{person}{Ross Wightman}, \bibinfo{person}{Mitchell
  Wortsman}, \bibinfo{person}{Gabriel Ilharco}, \bibinfo{person}{Cade Gordon},
  \bibinfo{person}{Christoph Schuhmann}, \bibinfo{person}{Ludwig Schmidt},
  {and} \bibinfo{person}{Jenia Jitsev}.} \bibinfo{year}{2023}\natexlab{}.
\newblock \showarticletitle{Reproducible scaling laws for contrastive
  language-image learning}. In \bibinfo{booktitle}{\emph{Proceedings of the
  IEEE/CVF Conference on Computer Vision and Pattern Recognition}}.
  \bibinfo{pages}{2818--2829}.
\newblock


\bibitem[Cong et~al\mbox{.}(2022)]%
        {cong2022satmae}
\bibfield{author}{\bibinfo{person}{Yezhen Cong}, \bibinfo{person}{Samar
  Khanna}, \bibinfo{person}{Chenlin Meng}, \bibinfo{person}{Patrick Liu},
  \bibinfo{person}{Erik Rozi}, \bibinfo{person}{Yutong He},
  \bibinfo{person}{Marshall Burke}, \bibinfo{person}{David Lobell}, {and}
  \bibinfo{person}{Stefano Ermon}.} \bibinfo{year}{2022}\natexlab{}.
\newblock \showarticletitle{Satmae: Pre-training transformers for temporal and
  multi-spectral satellite imagery}.
\newblock \bibinfo{journal}{\emph{Advances in Neural Information Processing
  Systems}}  \bibinfo{volume}{35} (\bibinfo{year}{2022}),
  \bibinfo{pages}{197--211}.
\newblock


\bibitem[Dai et~al\mbox{.}(2022)]%
        {dai2022enabling}
\bibfield{author}{\bibinfo{person}{Wenliang Dai}, \bibinfo{person}{Lu Hou},
  \bibinfo{person}{Lifeng Shang}, \bibinfo{person}{Xin Jiang},
  \bibinfo{person}{Qun Liu}, {and} \bibinfo{person}{Pascale Fung}.}
  \bibinfo{year}{2022}\natexlab{}.
\newblock \showarticletitle{Enabling multimodal generation on CLIP via
  vision-language knowledge distillation}.
\newblock \bibinfo{journal}{\emph{arXiv preprint arXiv:2203.06386}}
  (\bibinfo{year}{2022}).
\newblock


\bibitem[Ding et~al\mbox{.}(2023)]%
        {ding2023mgeo}
\bibfield{author}{\bibinfo{person}{Ruixue Ding}, \bibinfo{person}{Boli Chen},
  \bibinfo{person}{Pengjun Xie}, \bibinfo{person}{Fei Huang},
  \bibinfo{person}{Xin Li}, \bibinfo{person}{Qiang Zhang}, {and}
  \bibinfo{person}{Yao Xu}.} \bibinfo{year}{2023}\natexlab{}.
\newblock \showarticletitle{MGeo: Multi-Modal Geographic Language Model
  Pre-Training}. In \bibinfo{booktitle}{\emph{Proceedings of the 46th
  International ACM SIGIR Conference on Research and Development in Information
  Retrieval}}. \bibinfo{pages}{185--194}.
\newblock


\bibitem[Dong et~al\mbox{.}(2021)]%
        {dong2021attention}
\bibfield{author}{\bibinfo{person}{Yihe Dong}, \bibinfo{person}{Jean-Baptiste
  Cordonnier}, {and} \bibinfo{person}{Andreas Loukas}.}
  \bibinfo{year}{2021}\natexlab{}.
\newblock \showarticletitle{Attention is not all you need: Pure attention loses
  rank doubly exponentially with depth}. In
  \bibinfo{booktitle}{\emph{International Conference on Machine Learning}}.
  PMLR, \bibinfo{pages}{2793--2803}.
\newblock


\bibitem[Dosovitskiy et~al\mbox{.}(2020)]%
        {dosovitskiy2020image}
\bibfield{author}{\bibinfo{person}{Alexey Dosovitskiy}, \bibinfo{person}{Lucas
  Beyer}, \bibinfo{person}{Alexander Kolesnikov}, \bibinfo{person}{Dirk
  Weissenborn}, \bibinfo{person}{Xiaohua Zhai}, \bibinfo{person}{Thomas
  Unterthiner}, \bibinfo{person}{Mostafa Dehghani}, \bibinfo{person}{Matthias
  Minderer}, \bibinfo{person}{Georg Heigold}, \bibinfo{person}{Sylvain Gelly},
  {et~al\mbox{.}}} \bibinfo{year}{2020}\natexlab{}.
\newblock \showarticletitle{An image is worth 16x16 words: Transformers for
  image recognition at scale}.
\newblock \bibinfo{journal}{\emph{arXiv preprint arXiv:2010.11929}}
  (\bibinfo{year}{2020}).
\newblock


\bibitem[Gao et~al\mbox{.}(2023)]%
        {gao2023llama}
\bibfield{author}{\bibinfo{person}{Peng Gao}, \bibinfo{person}{Jiaming Han},
  \bibinfo{person}{Renrui Zhang}, \bibinfo{person}{Ziyi Lin},
  \bibinfo{person}{Shijie Geng}, \bibinfo{person}{Aojun Zhou},
  \bibinfo{person}{Wei Zhang}, \bibinfo{person}{Pan Lu},
  \bibinfo{person}{Conghui He}, \bibinfo{person}{Xiangyu Yue}, {et~al\mbox{.}}}
  \bibinfo{year}{2023}\natexlab{}.
\newblock \showarticletitle{Llama-adapter v2: Parameter-efficient visual
  instruction model}.
\newblock \bibinfo{journal}{\emph{arXiv preprint arXiv:2304.15010}}
  (\bibinfo{year}{2023}).
\newblock


\bibitem[Gao et~al\mbox{.}(2020)]%
        {gao2020making}
\bibfield{author}{\bibinfo{person}{Tianyu Gao}, \bibinfo{person}{Adam Fisch},
  {and} \bibinfo{person}{Danqi Chen}.} \bibinfo{year}{2020}\natexlab{}.
\newblock \showarticletitle{Making pre-trained language models better few-shot
  learners}.
\newblock \bibinfo{journal}{\emph{arXiv preprint arXiv:2012.15723}}
  (\bibinfo{year}{2020}).
\newblock


\bibitem[Gunasekar et~al\mbox{.}(2023)]%
        {gunasekar2023textbooks}
\bibfield{author}{\bibinfo{person}{Suriya Gunasekar}, \bibinfo{person}{Yi
  Zhang}, \bibinfo{person}{Jyoti Aneja}, \bibinfo{person}{Caio
  C{\'e}sar~Teodoro Mendes}, \bibinfo{person}{Allie Del~Giorno},
  \bibinfo{person}{Sivakanth Gopi}, \bibinfo{person}{Mojan Javaheripi},
  \bibinfo{person}{Piero Kauffmann}, \bibinfo{person}{Gustavo de Rosa},
  \bibinfo{person}{Olli Saarikivi}, {et~al\mbox{.}}}
  \bibinfo{year}{2023}\natexlab{}.
\newblock \showarticletitle{Textbooks Are All You Need}.
\newblock \bibinfo{journal}{\emph{arXiv preprint arXiv:2306.11644}}
  (\bibinfo{year}{2023}).
\newblock


\bibitem[Han et~al\mbox{.}(2023)]%
        {han2023imagebindllm}
\bibfield{author}{\bibinfo{person}{Jiaming Han}, \bibinfo{person}{Renrui
  Zhang}, \bibinfo{person}{Wenqi Shao}, \bibinfo{person}{Peng Gao},
  \bibinfo{person}{Peng Xu}, \bibinfo{person}{Han Xiao},
  \bibinfo{person}{Kaipeng Zhang}, \bibinfo{person}{Chris Liu},
  \bibinfo{person}{Song Wen}, \bibinfo{person}{Ziyu Guo},
  \bibinfo{person}{Xudong Lu}, \bibinfo{person}{Shuai Ren},
  \bibinfo{person}{Yafei Wen}, \bibinfo{person}{Xiaoxin Chen},
  \bibinfo{person}{Xiangyu Yue}, \bibinfo{person}{Hongsheng Li}, {and}
  \bibinfo{person}{Yu Qiao}.} \bibinfo{year}{2023}\natexlab{}.
\newblock \bibinfo{title}{ImageBind-LLM: Multi-modality Instruction Tuning}.
\newblock
\newblock
\showeprint[arxiv]{2309.03905}~[cs.MM]


\bibitem[Han et~al\mbox{.}(2022)]%
        {han2022survey}
\bibfield{author}{\bibinfo{person}{Kai Han}, \bibinfo{person}{Yunhe Wang},
  \bibinfo{person}{Hanting Chen}, \bibinfo{person}{Xinghao Chen},
  \bibinfo{person}{Jianyuan Guo}, \bibinfo{person}{Zhenhua Liu},
  \bibinfo{person}{Yehui Tang}, \bibinfo{person}{An Xiao},
  \bibinfo{person}{Chunjing Xu}, \bibinfo{person}{Yixing Xu}, {et~al\mbox{.}}}
  \bibinfo{year}{2022}\natexlab{}.
\newblock \showarticletitle{A survey on vision transformer}.
\newblock \bibinfo{journal}{\emph{IEEE transactions on pattern analysis and
  machine intelligence}} \bibinfo{volume}{45}, \bibinfo{number}{1}
  (\bibinfo{year}{2022}), \bibinfo{pages}{87--110}.
\newblock


\bibitem[Han et~al\mbox{.}(2020a)]%
        {han2020lightweight}
\bibfield{author}{\bibinfo{person}{Sungwon Han}, \bibinfo{person}{Donghyun
  Ahn}, \bibinfo{person}{Hyunji Cha}, \bibinfo{person}{Jeasurk Yang},
  \bibinfo{person}{Sungwon Park}, {and} \bibinfo{person}{Meeyoung Cha}.}
  \bibinfo{year}{2020}\natexlab{a}.
\newblock \showarticletitle{Lightweight and robust representation of economic
  scales from satellite imagery}. In \bibinfo{booktitle}{\emph{Proceedings of
  the AAAI Conference on Artificial Intelligence}}, Vol.~\bibinfo{volume}{34}.
  \bibinfo{pages}{428--436}.
\newblock


\bibitem[Han et~al\mbox{.}(2020b)]%
        {han2020learning}
\bibfield{author}{\bibinfo{person}{Sungwon Han}, \bibinfo{person}{Donghyun
  Ahn}, \bibinfo{person}{Sungwon Park}, \bibinfo{person}{Jeasurk Yang},
  \bibinfo{person}{Susang Lee}, \bibinfo{person}{Jihee Kim},
  \bibinfo{person}{Hyunjoo Yang}, \bibinfo{person}{Sangyoon Park}, {and}
  \bibinfo{person}{Meeyoung Cha}.} \bibinfo{year}{2020}\natexlab{b}.
\newblock \showarticletitle{Learning to score economic development from
  satellite imagery}. In \bibinfo{booktitle}{\emph{Proceedings of the 26th ACM
  SIGKDD International Conference on Knowledge Discovery \& Data Mining}}.
  \bibinfo{pages}{2970--2979}.
\newblock


\bibitem[Hao et~al\mbox{.}(2024)]%
        {hao2024urbanvlp}
\bibfield{author}{\bibinfo{person}{Xixuan Hao}, \bibinfo{person}{Wei Chen},
  \bibinfo{person}{Yibo Yan}, \bibinfo{person}{Siru Zhong},
  \bibinfo{person}{Kun Wang}, \bibinfo{person}{Qingsong Wen}, {and}
  \bibinfo{person}{Yuxuan Liang}.} \bibinfo{year}{2024}\natexlab{}.
\newblock \showarticletitle{UrbanVLP: A Multi-Granularity Vision-Language
  Pre-Trained Model for Urban Indicator Prediction}.
\newblock \bibinfo{journal}{\emph{arXiv preprint}} (\bibinfo{year}{2024}).
\newblock


\bibitem[He et~al\mbox{.}(2015)]%
        {he2015deep}
\bibfield{author}{\bibinfo{person}{Kaiming He}, \bibinfo{person}{Xiangyu
  Zhang}, \bibinfo{person}{Shaoqing Ren}, {and} \bibinfo{person}{Jian Sun}.}
  \bibinfo{year}{2015}\natexlab{}.
\newblock \bibinfo{title}{Deep Residual Learning for Image Recognition}.
\newblock
\newblock
\showeprint[arxiv]{1512.03385}~[cs.CV]


\bibitem[He et~al\mbox{.}(2018)]%
        {he2018perceiving}
\bibfield{author}{\bibinfo{person}{Zhiyuan He}, \bibinfo{person}{Su Yang},
  \bibinfo{person}{Weishan Zhang}, {and} \bibinfo{person}{Jiulong Zhang}.}
  \bibinfo{year}{2018}\natexlab{}.
\newblock \showarticletitle{Perceiving commerial activeness over satellite
  images}. In \bibinfo{booktitle}{\emph{Companion Proceedings of the The Web
  Conference 2018}}. \bibinfo{pages}{387--394}.
\newblock


\bibitem[Hong et~al\mbox{.}(2020)]%
        {hong2020graph}
\bibfield{author}{\bibinfo{person}{Danfeng Hong}, \bibinfo{person}{Lianru Gao},
  \bibinfo{person}{Jing Yao}, \bibinfo{person}{Bing Zhang},
  \bibinfo{person}{Antonio Plaza}, {and} \bibinfo{person}{Jocelyn Chanussot}.}
  \bibinfo{year}{2020}\natexlab{}.
\newblock \showarticletitle{Graph convolutional networks for hyperspectral
  image classification}.
\newblock \bibinfo{journal}{\emph{IEEE Transactions on Geoscience and Remote
  Sensing}} \bibinfo{volume}{59}, \bibinfo{number}{7} (\bibinfo{year}{2020}),
  \bibinfo{pages}{5966--5978}.
\newblock


\bibitem[Hu et~al\mbox{.}(2023)]%
        {hu2023survey}
\bibfield{author}{\bibinfo{person}{Linmei Hu}, \bibinfo{person}{Zeyi Liu},
  \bibinfo{person}{Ziwang Zhao}, \bibinfo{person}{Lei Hou},
  \bibinfo{person}{Liqiang Nie}, {and} \bibinfo{person}{Juanzi Li}.}
  \bibinfo{year}{2023}\natexlab{}.
\newblock \showarticletitle{A Survey of Knowledge Enhanced Pre-Trained Language
  Models}.
\newblock \bibinfo{journal}{\emph{IEEE Transactions on Knowledge and Data
  Engineering}} (\bibinfo{year}{2023}).
\newblock


\bibitem[Huang et~al\mbox{.}(2022)]%
        {huang2022ernie}
\bibfield{author}{\bibinfo{person}{Jizhou Huang}, \bibinfo{person}{Haifeng
  Wang}, \bibinfo{person}{Yibo Sun}, \bibinfo{person}{Yunsheng Shi},
  \bibinfo{person}{Zhengjie Huang}, \bibinfo{person}{An Zhuo}, {and}
  \bibinfo{person}{Shikun Feng}.} \bibinfo{year}{2022}\natexlab{}.
\newblock \showarticletitle{ERNIE-GeoL: A Geography-and-Language Pre-trained
  Model and its Applications in Baidu Maps}. In
  \bibinfo{booktitle}{\emph{Proceedings of the 28th ACM SIGKDD Conference on
  Knowledge Discovery and Data Mining}}. \bibinfo{pages}{3029--3039}.
\newblock


\bibitem[Huang et~al\mbox{.}(2021)]%
        {huang2021m3g}
\bibfield{author}{\bibinfo{person}{Tianyuan Huang}, \bibinfo{person}{Zhecheng
  Wang}, \bibinfo{person}{Hao Sheng}, \bibinfo{person}{Andrew~Y Ng}, {and}
  \bibinfo{person}{Ram Rajagopal}.} \bibinfo{year}{2021}\natexlab{}.
\newblock \showarticletitle{M3G: Learning urban neighborhood representation
  from multi-modal multi-graph}. In \bibinfo{booktitle}{\emph{Proceedings of
  the DeepSpatial 2021: 2nd ACM KDD Workshop on Deep Learning for
  Spatio-Temporal Data, Applications and Systems}}.
\newblock


\bibitem[Ilharco et~al\mbox{.}(2021)]%
        {ilharco_gabriel_2021_5143773}
\bibfield{author}{\bibinfo{person}{Gabriel Ilharco}, \bibinfo{person}{Mitchell
  Wortsman}, \bibinfo{person}{Ross Wightman}, \bibinfo{person}{Cade Gordon},
  \bibinfo{person}{Nicholas Carlini}, \bibinfo{person}{Rohan Taori},
  \bibinfo{person}{Achal Dave}, \bibinfo{person}{Vaishaal Shankar},
  \bibinfo{person}{Hongseok Namkoong}, \bibinfo{person}{John Miller},
  \bibinfo{person}{Hannaneh Hajishirzi}, \bibinfo{person}{Ali Farhadi}, {and}
  \bibinfo{person}{Ludwig Schmidt}.} \bibinfo{year}{2021}\natexlab{}.
\newblock \bibinfo{booktitle}{\emph{OpenCLIP}}.
\newblock
\urldef\tempurl%
\url{https://doi.org/10.5281/zenodo.5143773}
\showDOI{\tempurl}


\bibitem[Jean et~al\mbox{.}(2016)]%
        {jean2016combining}
\bibfield{author}{\bibinfo{person}{Neal Jean}, \bibinfo{person}{Marshall
  Burke}, \bibinfo{person}{Michael Xie}, \bibinfo{person}{W~Matthew Davis},
  \bibinfo{person}{David~B Lobell}, {and} \bibinfo{person}{Stefano Ermon}.}
  \bibinfo{year}{2016}\natexlab{}.
\newblock \showarticletitle{Combining satellite imagery and machine learning to
  predict poverty}.
\newblock \bibinfo{journal}{\emph{Science}} \bibinfo{volume}{353},
  \bibinfo{number}{6301} (\bibinfo{year}{2016}), \bibinfo{pages}{790--794}.
\newblock


\bibitem[Jean et~al\mbox{.}(2019)]%
        {jean2019tile2vec}
\bibfield{author}{\bibinfo{person}{Neal Jean}, \bibinfo{person}{Sherrie Wang},
  \bibinfo{person}{Anshul Samar}, \bibinfo{person}{George Azzari},
  \bibinfo{person}{David Lobell}, {and} \bibinfo{person}{Stefano Ermon}.}
  \bibinfo{year}{2019}\natexlab{}.
\newblock \showarticletitle{Tile2vec: Unsupervised representation learning for
  spatially distributed data}. In \bibinfo{booktitle}{\emph{Proceedings of the
  AAAI Conference on Artificial Intelligence}}, Vol.~\bibinfo{volume}{33}.
  \bibinfo{pages}{3967--3974}.
\newblock


\bibitem[Jenkins et~al\mbox{.}(2019)]%
        {jenkins2019unsupervised}
\bibfield{author}{\bibinfo{person}{Porter Jenkins}, \bibinfo{person}{Ahmad
  Farag}, \bibinfo{person}{Suhang Wang}, {and} \bibinfo{person}{Zhenhui Li}.}
  \bibinfo{year}{2019}\natexlab{}.
\newblock \showarticletitle{Unsupervised representation learning of spatial
  data via multimodal embedding}. In \bibinfo{booktitle}{\emph{Proceedings of
  the 28th ACM international conference on information and knowledge
  management}}. \bibinfo{pages}{1993--2002}.
\newblock


\bibitem[Jiang et~al\mbox{.}(2020)]%
        {jiang2020can}
\bibfield{author}{\bibinfo{person}{Zhengbao Jiang}, \bibinfo{person}{Frank~F
  Xu}, \bibinfo{person}{Jun Araki}, {and} \bibinfo{person}{Graham Neubig}.}
  \bibinfo{year}{2020}\natexlab{}.
\newblock \showarticletitle{How can we know what language models know?}
\newblock \bibinfo{journal}{\emph{Transactions of the Association for
  Computational Linguistics}}  \bibinfo{volume}{8} (\bibinfo{year}{2020}),
  \bibinfo{pages}{423--438}.
\newblock


\bibitem[Jin et~al\mbox{.}(2023a)]%
        {jintime2023}
\bibfield{author}{\bibinfo{person}{Ming Jin}, \bibinfo{person}{Shiyu Wang},
  \bibinfo{person}{Lintao Ma}, \bibinfo{person}{Zhixuan Chu},
  \bibinfo{person}{James~Y. Zhang}, \bibinfo{person}{Xiaoming Shi},
  \bibinfo{person}{Pin-Yu Chen}, \bibinfo{person}{Yuxuan Liang},
  \bibinfo{person}{Yuan-Fang Li}, \bibinfo{person}{Shirui Pan}, {and}
  \bibinfo{person}{Qingsong Wen}.} \bibinfo{year}{2023}\natexlab{a}.
\newblock \showarticletitle{{Time-LLM}: Time Series Forecasting by
  Reprogramming Large Language Models}.
\newblock \bibinfo{journal}{\emph{arXiv preprint arXiv:2310.01728}}
  (\bibinfo{year}{2023}).
\newblock


\bibitem[Jin et~al\mbox{.}(2023b)]%
        {jin2023timellm}
\bibfield{author}{\bibinfo{person}{Ming Jin}, \bibinfo{person}{Shiyu Wang},
  \bibinfo{person}{Lintao Ma}, \bibinfo{person}{Zhixuan Chu},
  \bibinfo{person}{James~Y. Zhang}, \bibinfo{person}{Xiaoming Shi},
  \bibinfo{person}{Pin-Yu Chen}, \bibinfo{person}{Yuxuan Liang},
  \bibinfo{person}{Yuan-Fang Li}, \bibinfo{person}{Shirui Pan}, {and}
  \bibinfo{person}{Qingsong Wen}.} \bibinfo{year}{2023}\natexlab{b}.
\newblock \showarticletitle{{Time-LLM}: Time Series Forecasting by
  Reprogramming Large Language Models}.
\newblock \bibinfo{journal}{\emph{arXiv preprint arXiv:2310.01728}}
  (\bibinfo{year}{2023}).
\newblock


\bibitem[Kang et~al\mbox{.}(2020)]%
        {kang2020deep}
\bibfield{author}{\bibinfo{person}{Jian Kang}, \bibinfo{person}{Ruben
  Fernandez-Beltran}, \bibinfo{person}{Puhong Duan}, \bibinfo{person}{Sicong
  Liu}, {and} \bibinfo{person}{Antonio~J Plaza}.}
  \bibinfo{year}{2020}\natexlab{}.
\newblock \showarticletitle{Deep unsupervised embedding for remotely sensed
  images based on spatially augmented momentum contrast}.
\newblock \bibinfo{journal}{\emph{IEEE Transactions on Geoscience and Remote
  Sensing}} \bibinfo{volume}{59}, \bibinfo{number}{3} (\bibinfo{year}{2020}),
  \bibinfo{pages}{2598--2610}.
\newblock


\bibitem[Kenton and Toutanova(2019)]%
        {kenton2019bert}
\bibfield{author}{\bibinfo{person}{Jacob Devlin Ming-Wei~Chang Kenton} {and}
  \bibinfo{person}{Lee~Kristina Toutanova}.} \bibinfo{year}{2019}\natexlab{}.
\newblock \showarticletitle{BERT: Pre-training of Deep Bidirectional
  Transformers for Language Understanding}. In
  \bibinfo{booktitle}{\emph{Proceedings of NAACL-HLT}}.
  \bibinfo{pages}{4171--4186}.
\newblock


\bibitem[Kim et~al\mbox{.}(2021)]%
        {kim2021vilt}
\bibfield{author}{\bibinfo{person}{Wonjae Kim}, \bibinfo{person}{Bokyung Son},
  {and} \bibinfo{person}{Ildoo Kim}.} \bibinfo{year}{2021}\natexlab{}.
\newblock \showarticletitle{Vilt: Vision-and-language transformer without
  convolution or region supervision}. In
  \bibinfo{booktitle}{\emph{International Conference on Machine Learning}}.
  PMLR, \bibinfo{pages}{5583--5594}.
\newblock


\bibitem[Kramer(1991)]%
        {kramer1991nonlinear}
\bibfield{author}{\bibinfo{person}{Mark~A Kramer}.}
  \bibinfo{year}{1991}\natexlab{}.
\newblock \showarticletitle{Nonlinear principal component analysis using
  autoassociative neural networks}.
\newblock \bibinfo{journal}{\emph{AIChE journal}} \bibinfo{volume}{37},
  \bibinfo{number}{2} (\bibinfo{year}{1991}), \bibinfo{pages}{233--243}.
\newblock


\bibitem[Law et~al\mbox{.}(2019)]%
        {law2019take}
\bibfield{author}{\bibinfo{person}{Stephen Law}, \bibinfo{person}{Brooks
  Paige}, {and} \bibinfo{person}{Chris Russell}.}
  \bibinfo{year}{2019}\natexlab{}.
\newblock \showarticletitle{Take a look around: using street view and satellite
  images to estimate house prices}.
\newblock \bibinfo{journal}{\emph{ACM Transactions on Intelligent Systems and
  Technology (TIST)}} \bibinfo{volume}{10}, \bibinfo{number}{5}
  (\bibinfo{year}{2019}), \bibinfo{pages}{1--19}.
\newblock


\bibitem[Lee et~al\mbox{.}(2019)]%
        {lee2019set}
\bibfield{author}{\bibinfo{person}{Juho Lee}, \bibinfo{person}{Yoonho Lee},
  \bibinfo{person}{Jungtaek Kim}, \bibinfo{person}{Adam Kosiorek},
  \bibinfo{person}{Seungjin Choi}, {and} \bibinfo{person}{Yee~Whye Teh}.}
  \bibinfo{year}{2019}\natexlab{}.
\newblock \showarticletitle{Set transformer: A framework for attention-based
  permutation-invariant neural networks}. In
  \bibinfo{booktitle}{\emph{International conference on machine learning}}.
  PMLR, \bibinfo{pages}{3744--3753}.
\newblock


\bibitem[Lester et~al\mbox{.}(2021)]%
        {lester2021power}
\bibfield{author}{\bibinfo{person}{Brian Lester}, \bibinfo{person}{Rami
  Al-Rfou}, {and} \bibinfo{person}{Noah Constant}.}
  \bibinfo{year}{2021}\natexlab{}.
\newblock \showarticletitle{The power of scale for parameter-efficient prompt
  tuning}.
\newblock \bibinfo{journal}{\emph{arXiv preprint arXiv:2104.08691}}
  (\bibinfo{year}{2021}).
\newblock


\bibitem[Li et~al\mbox{.}(2022e)]%
        {li-etal-2022-mplug}
\bibfield{author}{\bibinfo{person}{Chenliang Li}, \bibinfo{person}{Haiyang Xu},
  \bibinfo{person}{Junfeng Tian}, \bibinfo{person}{Wei Wang},
  \bibinfo{person}{Ming Yan}, \bibinfo{person}{Bin Bi}, \bibinfo{person}{Jiabo
  Ye}, \bibinfo{person}{He Chen}, \bibinfo{person}{Guohai Xu},
  \bibinfo{person}{Zheng Cao}, \bibinfo{person}{Ji Zhang},
  \bibinfo{person}{Songfang Huang}, \bibinfo{person}{Fei Huang},
  \bibinfo{person}{Jingren Zhou}, {and} \bibinfo{person}{Luo Si}.}
  \bibinfo{year}{2022}\natexlab{e}.
\newblock \showarticletitle{m{PLUG}: Effective and Efficient Vision-Language
  Learning by Cross-modal Skip-connections}. In
  \bibinfo{booktitle}{\emph{Proceedings of the 2022 Conference on Empirical
  Methods in Natural Language Processing}}. \bibinfo{publisher}{Association for
  Computational Linguistics}, \bibinfo{address}{Abu Dhabi, United Arab
  Emirates}, \bibinfo{pages}{7241--7259}.
\newblock
\urldef\tempurl%
\url{https://doi.org/10.18653/v1/2022.emnlp-main.488}
\showDOI{\tempurl}


\bibitem[Li et~al\mbox{.}(2023)]%
        {li2023blip}
\bibfield{author}{\bibinfo{person}{Junnan Li}, \bibinfo{person}{Dongxu Li},
  \bibinfo{person}{Silvio Savarese}, {and} \bibinfo{person}{Steven Hoi}.}
  \bibinfo{year}{2023}\natexlab{}.
\newblock \showarticletitle{Blip-2: Bootstrapping language-image pre-training
  with frozen image encoders and large language models}.
\newblock \bibinfo{journal}{\emph{arXiv preprint arXiv:2301.12597}}
  (\bibinfo{year}{2023}).
\newblock


\bibitem[Li et~al\mbox{.}(2022b)]%
        {li2022blip}
\bibfield{author}{\bibinfo{person}{Junnan Li}, \bibinfo{person}{Dongxu Li},
  \bibinfo{person}{Caiming Xiong}, {and} \bibinfo{person}{Steven Hoi}.}
  \bibinfo{year}{2022}\natexlab{b}.
\newblock \showarticletitle{Blip: Bootstrapping language-image pre-training for
  unified vision-language understanding and generation}. In
  \bibinfo{booktitle}{\emph{International Conference on Machine Learning}}.
  PMLR, \bibinfo{pages}{12888--12900}.
\newblock


\bibitem[Li et~al\mbox{.}(2021)]%
        {li2021align}
\bibfield{author}{\bibinfo{person}{Junnan Li}, \bibinfo{person}{Ramprasaath
  Selvaraju}, \bibinfo{person}{Akhilesh Gotmare}, \bibinfo{person}{Shafiq
  Joty}, \bibinfo{person}{Caiming Xiong}, {and} \bibinfo{person}{Steven
  Chu~Hong Hoi}.} \bibinfo{year}{2021}\natexlab{}.
\newblock \showarticletitle{Align before fuse: Vision and language
  representation learning with momentum distillation}.
\newblock \bibinfo{journal}{\emph{Advances in neural information processing
  systems}}  \bibinfo{volume}{34} (\bibinfo{year}{2021}),
  \bibinfo{pages}{9694--9705}.
\newblock


\bibitem[Li et~al\mbox{.}(2022c)]%
        {li2022pretrained}
\bibfield{author}{\bibinfo{person}{Junyi Li}, \bibinfo{person}{Tianyi Tang},
  \bibinfo{person}{Wayne~Xin Zhao}, \bibinfo{person}{Jian-Yun Nie}, {and}
  \bibinfo{person}{Ji-Rong Wen}.} \bibinfo{year}{2022}\natexlab{c}.
\newblock \showarticletitle{Pretrained language models for text generation: A
  survey}.
\newblock \bibinfo{journal}{\emph{arXiv preprint arXiv:2201.05273}}
  (\bibinfo{year}{2022}).
\newblock


\bibitem[Li et~al\mbox{.}(2022d)]%
        {li2022predicting}
\bibfield{author}{\bibinfo{person}{Tong Li}, \bibinfo{person}{Shiduo Xin},
  \bibinfo{person}{Yanxin Xi}, \bibinfo{person}{Sasu Tarkoma},
  \bibinfo{person}{Pan Hui}, {and} \bibinfo{person}{Yong Li}.}
  \bibinfo{year}{2022}\natexlab{d}.
\newblock \showarticletitle{Predicting multi-level socioeconomic indicators
  from structural urban imagery}. In \bibinfo{booktitle}{\emph{Proceedings of
  the 31st ACM International Conference on Information \& Knowledge
  Management}}. \bibinfo{pages}{3282--3291}.
\newblock


\bibitem[Li and Liang(2021)]%
        {li2021prefix}
\bibfield{author}{\bibinfo{person}{Xiang~Lisa Li} {and} \bibinfo{person}{Percy
  Liang}.} \bibinfo{year}{2021}\natexlab{}.
\newblock \showarticletitle{Prefix-tuning: Optimizing continuous prompts for
  generation}.
\newblock \bibinfo{journal}{\emph{arXiv preprint arXiv:2101.00190}}
  (\bibinfo{year}{2021}).
\newblock


\bibitem[Li et~al\mbox{.}(2022a)]%
        {li2022competition}
\bibfield{author}{\bibinfo{person}{Yujia Li}, \bibinfo{person}{David Choi},
  \bibinfo{person}{Junyoung Chung}, \bibinfo{person}{Nate Kushman},
  \bibinfo{person}{Julian Schrittwieser}, \bibinfo{person}{R{\'e}mi Leblond},
  \bibinfo{person}{Tom Eccles}, \bibinfo{person}{James Keeling},
  \bibinfo{person}{Felix Gimeno}, \bibinfo{person}{Agustin Dal~Lago},
  {et~al\mbox{.}}} \bibinfo{year}{2022}\natexlab{a}.
\newblock \showarticletitle{Competition-level code generation with alphacode}.
\newblock \bibinfo{journal}{\emph{Science}} \bibinfo{volume}{378},
  \bibinfo{number}{6624} (\bibinfo{year}{2022}), \bibinfo{pages}{1092--1097}.
\newblock


\bibitem[Liu et~al\mbox{.}(2023a)]%
        {liu2023llava}
\bibfield{author}{\bibinfo{person}{Haotian Liu}, \bibinfo{person}{Chunyuan Li},
  \bibinfo{person}{Qingyang Wu}, {and} \bibinfo{person}{Yong~Jae Lee}.}
  \bibinfo{year}{2023}\natexlab{a}.
\newblock \bibinfo{title}{Visual Instruction Tuning}.
\newblock
\newblock


\bibitem[Liu et~al\mbox{.}(2023b)]%
        {liu2023pre}
\bibfield{author}{\bibinfo{person}{Pengfei Liu}, \bibinfo{person}{Weizhe Yuan},
  \bibinfo{person}{Jinlan Fu}, \bibinfo{person}{Zhengbao Jiang},
  \bibinfo{person}{Hiroaki Hayashi}, {and} \bibinfo{person}{Graham Neubig}.}
  \bibinfo{year}{2023}\natexlab{b}.
\newblock \showarticletitle{Pre-train, prompt, and predict: A systematic survey
  of prompting methods in natural language processing}.
\newblock \bibinfo{journal}{\emph{Comput. Surveys}} \bibinfo{volume}{55},
  \bibinfo{number}{9} (\bibinfo{year}{2023}), \bibinfo{pages}{1--35}.
\newblock


\bibitem[Liu et~al\mbox{.}(2023c)]%
        {liu2023knowledge}
\bibfield{author}{\bibinfo{person}{Yu Liu}, \bibinfo{person}{Xin Zhang},
  \bibinfo{person}{Jingtao Ding}, \bibinfo{person}{Yanxin Xi}, {and}
  \bibinfo{person}{Yong Li}.} \bibinfo{year}{2023}\natexlab{c}.
\newblock \showarticletitle{Knowledge-infused contrastive learning for urban
  imagery-based socioeconomic prediction}. In
  \bibinfo{booktitle}{\emph{Proceedings of the ACM Web Conference 2023}}.
  \bibinfo{pages}{4150--4160}.
\newblock


\bibitem[M~Rustowicz et~al\mbox{.}(2019)]%
        {m2019semantic}
\bibfield{author}{\bibinfo{person}{Rose M~Rustowicz}, \bibinfo{person}{Robin
  Cheong}, \bibinfo{person}{Lijing Wang}, \bibinfo{person}{Stefano Ermon},
  \bibinfo{person}{Marshall Burke}, {and} \bibinfo{person}{David Lobell}.}
  \bibinfo{year}{2019}\natexlab{}.
\newblock \showarticletitle{Semantic segmentation of crop type in Africa: A
  novel dataset and analysis of deep learning methods}. In
  \bibinfo{booktitle}{\emph{Proceedings of the IEEE/CVF Conference on Computer
  Vision and Pattern Recognition Workshops}}. \bibinfo{pages}{75--82}.
\newblock


\bibitem[Manvi et~al\mbox{.}(2023)]%
        {manvi2023geollm}
\bibfield{author}{\bibinfo{person}{Rohin Manvi}, \bibinfo{person}{Samar
  Khanna}, \bibinfo{person}{Gengchen Mai}, \bibinfo{person}{Marshall Burke},
  \bibinfo{person}{David Lobell}, {and} \bibinfo{person}{Stefano Ermon}.}
  \bibinfo{year}{2023}\natexlab{}.
\newblock \showarticletitle{Geollm: Extracting geospatial knowledge from large
  language models}.
\newblock \bibinfo{journal}{\emph{arXiv preprint arXiv:2310.06213}}
  (\bibinfo{year}{2023}).
\newblock


\bibitem[{Mapbox}({[n.\,d.]})]%
        {mapbox}
\bibfield{author}{\bibinfo{person}{{Mapbox}}.}
  \bibinfo{year}{[n.\,d.]}\natexlab{}.
\newblock \bibinfo{booktitle}{\emph{{Mapbox - Location Data \& Maps for
  Developers}}}.
\newblock
\urldef\tempurl%
\url{https://www.mapbox.com/}
\showURL{%
\tempurl}


\bibitem[Martinez et~al\mbox{.}(2021)]%
        {martinez2021fully}
\bibfield{author}{\bibinfo{person}{Jorge Andres~Chamorro Martinez},
  \bibinfo{person}{Laura Elena~Cu{\'e} La~Rosa}, \bibinfo{person}{Raul~Queiroz
  Feitosa}, \bibinfo{person}{Ieda~Del’Arco Sanches}, {and}
  \bibinfo{person}{Patrick~Nigri Happ}.} \bibinfo{year}{2021}\natexlab{}.
\newblock \showarticletitle{Fully convolutional recurrent networks for
  multidate crop recognition from multitemporal image sequences}.
\newblock \bibinfo{journal}{\emph{ISPRS Journal of Photogrammetry and Remote
  Sensing}}  \bibinfo{volume}{171} (\bibinfo{year}{2021}),
  \bibinfo{pages}{188--201}.
\newblock


\bibitem[{MEASURE DHS et al.}(2013)]%
        {measureDHS2013}
\bibfield{author}{\bibinfo{person}{{MEASURE DHS et al.}}}
  \bibinfo{year}{2013}\natexlab{}.
\newblock \bibinfo{booktitle}{\emph{Demographic and Health Surveys}}.
\newblock \bibinfo{publisher}{Measure DHS}, \bibinfo{address}{Calverton}.
\newblock


\bibitem[Miller(2004)]%
        {miller2004tobler}
\bibfield{author}{\bibinfo{person}{Harvey~J Miller}.}
  \bibinfo{year}{2004}\natexlab{}.
\newblock \showarticletitle{Tobler's first law and spatial analysis}.
\newblock \bibinfo{journal}{\emph{Annals of the association of American
  geographers}} \bibinfo{volume}{94}, \bibinfo{number}{2}
  (\bibinfo{year}{2004}), \bibinfo{pages}{284--289}.
\newblock


\bibitem[Naizhuo~Zhao and Zhang(2017)]%
        {doi:10.1080/15481603.2016.1276705}
\bibfield{author}{\bibinfo{person}{Guofeng Cao Eric L.~Samson Naizhuo~Zhao,
  Ying~Liu} {and} \bibinfo{person}{Jingqi Zhang}.}
  \bibinfo{year}{2017}\natexlab{}.
\newblock \showarticletitle{Forecasting China’s GDP at the pixel level using
  nighttime lights time series and population images}.
\newblock \bibinfo{journal}{\emph{GIScience \& Remote Sensing}}
  \bibinfo{volume}{54}, \bibinfo{number}{3} (\bibinfo{year}{2017}),
  \bibinfo{pages}{407--425}.
\newblock
\urldef\tempurl%
\url{https://doi.org/10.1080/15481603.2016.1276705}
\showDOI{\tempurl}


\bibitem[Oda et~al\mbox{.}(2018)]%
        {oda2018open}
\bibfield{author}{\bibinfo{person}{Tomohiro Oda}, \bibinfo{person}{Shamil
  Maksyutov}, {and} \bibinfo{person}{Robert~J Andres}.}
  \bibinfo{year}{2018}\natexlab{}.
\newblock \showarticletitle{The Open-source Data Inventory for Anthropogenic CO
  2, version 2016 (ODIAC2016): a global monthly fossil fuel CO 2 gridded
  emissions data product for tracer transport simulations and surface flux
  inversions}.
\newblock \bibinfo{journal}{\emph{Earth System Science Data}}
  \bibinfo{volume}{10}, \bibinfo{number}{1} (\bibinfo{year}{2018}),
  \bibinfo{pages}{87--107}.
\newblock


\bibitem[Pan et~al\mbox{.}(2023)]%
        {pan2023retrieving}
\bibfield{author}{\bibinfo{person}{Junting Pan}, \bibinfo{person}{Ziyi Lin},
  \bibinfo{person}{Yuying Ge}, \bibinfo{person}{Xiatian Zhu},
  \bibinfo{person}{Renrui Zhang}, \bibinfo{person}{Yi Wang},
  \bibinfo{person}{Yu Qiao}, {and} \bibinfo{person}{Hongsheng Li}.}
  \bibinfo{year}{2023}\natexlab{}.
\newblock \showarticletitle{Retrieving-to-Answer: Zero-Shot Video Question
  Answering with Frozen Large Language Models}.
\newblock \bibinfo{journal}{\emph{arXiv preprint arXiv:2306.11732}}
  (\bibinfo{year}{2023}).
\newblock


\bibitem[Park et~al\mbox{.}(2022)]%
        {park2022learning}
\bibfield{author}{\bibinfo{person}{Sungwon Park}, \bibinfo{person}{Sungwon
  Han}, \bibinfo{person}{Donghyun Ahn}, \bibinfo{person}{Jaeyeon Kim},
  \bibinfo{person}{Jeasurk Yang}, \bibinfo{person}{Susang Lee},
  \bibinfo{person}{Seunghoon Hong}, \bibinfo{person}{Jihee Kim},
  \bibinfo{person}{Sangyoon Park}, \bibinfo{person}{Hyunjoo Yang},
  {et~al\mbox{.}}} \bibinfo{year}{2022}\natexlab{}.
\newblock \showarticletitle{Learning economic indicators by aggregating
  multi-level geospatial information}. In \bibinfo{booktitle}{\emph{Proceedings
  of the AAAI Conference on Artificial Intelligence}},
  Vol.~\bibinfo{volume}{36}. \bibinfo{pages}{12053--12061}.
\newblock


\bibitem[Perez et~al\mbox{.}(2017)]%
        {perez2017poverty}
\bibfield{author}{\bibinfo{person}{Anthony Perez}, \bibinfo{person}{Christopher
  Yeh}, \bibinfo{person}{George Azzari}, \bibinfo{person}{Marshall Burke},
  \bibinfo{person}{David Lobell}, {and} \bibinfo{person}{Stefano Ermon}.}
  \bibinfo{year}{2017}\natexlab{}.
\newblock \showarticletitle{Poverty prediction with public landsat 7 satellite
  imagery and machine learning}.
\newblock \bibinfo{journal}{\emph{arXiv preprint arXiv:1711.03654}}
  (\bibinfo{year}{2017}).
\newblock


\bibitem[Radford et~al\mbox{.}(2021)]%
        {radford2021learning}
\bibfield{author}{\bibinfo{person}{Alec Radford}, \bibinfo{person}{Jong~Wook
  Kim}, \bibinfo{person}{Chris Hallacy}, \bibinfo{person}{Aditya Ramesh},
  \bibinfo{person}{Gabriel Goh}, \bibinfo{person}{Sandhini Agarwal},
  \bibinfo{person}{Girish Sastry}, \bibinfo{person}{Amanda Askell},
  \bibinfo{person}{Pamela Mishkin}, \bibinfo{person}{Jack Clark},
  {et~al\mbox{.}}} \bibinfo{year}{2021}\natexlab{}.
\newblock \showarticletitle{Learning transferable visual models from natural
  language supervision}. In \bibinfo{booktitle}{\emph{International conference
  on machine learning}}. PMLR, \bibinfo{pages}{8748--8763}.
\newblock


\bibitem[Ritchie and Roser(2018)]%
        {owidurbanization}
\bibfield{author}{\bibinfo{person}{Hannah Ritchie} {and} \bibinfo{person}{Max
  Roser}.} \bibinfo{year}{2018}\natexlab{}.
\newblock \showarticletitle{Urbanization}.
\newblock \bibinfo{journal}{\emph{Our World in Data}} (\bibinfo{year}{2018}).
\newblock
\newblock
\shownote{https://ourworldindata.org/urbanization}.


\bibitem[Ru{\ss}wurm and K{\"o}rner(2020)]%
        {russwurm2020self}
\bibfield{author}{\bibinfo{person}{Marc Ru{\ss}wurm} {and}
  \bibinfo{person}{Marco K{\"o}rner}.} \bibinfo{year}{2020}\natexlab{}.
\newblock \showarticletitle{Self-attention for raw optical satellite time
  series classification}.
\newblock \bibinfo{journal}{\emph{ISPRS journal of photogrammetry and remote
  sensing}}  \bibinfo{volume}{169} (\bibinfo{year}{2020}),
  \bibinfo{pages}{421--435}.
\newblock


\bibitem[Shuster et~al\mbox{.}(2022)]%
        {shuster2022blenderbot}
\bibfield{author}{\bibinfo{person}{Kurt Shuster}, \bibinfo{person}{Jing Xu},
  \bibinfo{person}{Mojtaba Komeili}, \bibinfo{person}{Da Ju},
  \bibinfo{person}{Eric~Michael Smith}, \bibinfo{person}{Stephen Roller},
  \bibinfo{person}{Megan Ung}, \bibinfo{person}{Moya Chen},
  \bibinfo{person}{Kushal Arora}, \bibinfo{person}{Joshua Lane},
  {et~al\mbox{.}}} \bibinfo{year}{2022}\natexlab{}.
\newblock \showarticletitle{Blenderbot 3: a deployed conversational agent that
  continually learns to responsibly engage}.
\newblock \bibinfo{journal}{\emph{arXiv preprint arXiv:2208.03188}}
  (\bibinfo{year}{2022}).
\newblock


\bibitem[Su et~al\mbox{.}(2023b)]%
        {su2023language}
\bibfield{author}{\bibinfo{person}{Hung-Ting Su}, \bibinfo{person}{Yulei Niu},
  \bibinfo{person}{Xudong Lin}, \bibinfo{person}{Winston~H Hsu}, {and}
  \bibinfo{person}{Shih-Fu Chang}.} \bibinfo{year}{2023}\natexlab{b}.
\newblock \showarticletitle{Language Models are Causal Knowledge Extractors for
  Zero-shot Video Question Answering}. In \bibinfo{booktitle}{\emph{Proceedings
  of the IEEE/CVF Conference on Computer Vision and Pattern Recognition}}.
  \bibinfo{pages}{4950--4959}.
\newblock


\bibitem[Su et~al\mbox{.}(2023a)]%
        {su2023pandagpt}
\bibfield{author}{\bibinfo{person}{Yixuan Su}, \bibinfo{person}{Tian Lan},
  \bibinfo{person}{Huayang Li}, \bibinfo{person}{Jialu Xu},
  \bibinfo{person}{Yan Wang}, {and} \bibinfo{person}{Deng Cai}.}
  \bibinfo{year}{2023}\natexlab{a}.
\newblock \showarticletitle{Pandagpt: One model to instruction-follow them
  all}.
\newblock \bibinfo{journal}{\emph{arXiv preprint arXiv:2305.16355}}
  (\bibinfo{year}{2023}).
\newblock


\bibitem[Sultan et~al\mbox{.}(2023)]%
        {sultan2023geosam}
\bibfield{author}{\bibinfo{person}{Rafi~Ibn Sultan}, \bibinfo{person}{Chengyin
  Li}, \bibinfo{person}{Hui Zhu}, \bibinfo{person}{Prashant Khanduri},
  \bibinfo{person}{Marco Brocanelli}, {and} \bibinfo{person}{Dongxiao Zhu}.}
  \bibinfo{year}{2023}\natexlab{}.
\newblock \showarticletitle{GeoSAM: Fine-tuning SAM with sparse and dense
  visual prompting for automated segmentation of mobility infrastructure}.
\newblock \bibinfo{journal}{\emph{arXiv preprint arXiv:2311.11319}}
  (\bibinfo{year}{2023}).
\newblock


\bibitem[Sun et~al\mbox{.}(2023)]%
        {sun2023generative}
\bibfield{author}{\bibinfo{person}{Quan Sun}, \bibinfo{person}{Qiying Yu},
  \bibinfo{person}{Yufeng Cui}, \bibinfo{person}{Fan Zhang},
  \bibinfo{person}{Xiaosong Zhang}, \bibinfo{person}{Yueze Wang},
  \bibinfo{person}{Hongcheng Gao}, \bibinfo{person}{Jingjing Liu},
  \bibinfo{person}{Tiejun Huang}, {and} \bibinfo{person}{Xinlong Wang}.}
  \bibinfo{year}{2023}\natexlab{}.
\newblock \bibinfo{title}{Generative Pretraining in Multimodality}.
\newblock
\newblock
\showeprint[arxiv]{2307.05222}~[cs.CV]


\bibitem[Tang et~al\mbox{.}(2024)]%
        {tang2024synergizing}
\bibfield{author}{\bibinfo{person}{Yihong Tang}, \bibinfo{person}{Zhaokai
  Wang}, \bibinfo{person}{Ao Qu}, \bibinfo{person}{Yihao Yan},
  \bibinfo{person}{Kebing Hou}, \bibinfo{person}{Dingyi Zhuang},
  \bibinfo{person}{Xiaotong Guo}, \bibinfo{person}{Jinhua Zhao},
  \bibinfo{person}{Zhan Zhao}, {and} \bibinfo{person}{Wei Ma}.}
  \bibinfo{year}{2024}\natexlab{}.
\newblock \showarticletitle{Synergizing Spatial Optimization with Large
  Language Models for Open-Domain Urban Itinerary Planning}.
\newblock \bibinfo{journal}{\emph{arXiv preprint arXiv:2402.07204}}
  (\bibinfo{year}{2024}).
\newblock


\bibitem[Tatem(2017)]%
        {worldpop}
\bibfield{author}{\bibinfo{person}{Andrew~J Tatem}.}
  \bibinfo{year}{2017}\natexlab{}.
\newblock \showarticletitle{WorldPop, open data for spatial demography}.
\newblock \bibinfo{journal}{\emph{Scientific data}} \bibinfo{volume}{4},
  \bibinfo{number}{1} (\bibinfo{year}{2017}), \bibinfo{pages}{1--4}.
\newblock


\bibitem[Tewel et~al\mbox{.}(2022)]%
        {tewel2022zerocap}
\bibfield{author}{\bibinfo{person}{Yoad Tewel}, \bibinfo{person}{Yoav Shalev},
  \bibinfo{person}{Idan Schwartz}, {and} \bibinfo{person}{Lior Wolf}.}
  \bibinfo{year}{2022}\natexlab{}.
\newblock \showarticletitle{Zerocap: Zero-shot image-to-text generation for
  visual-semantic arithmetic}. In \bibinfo{booktitle}{\emph{Proceedings of the
  IEEE/CVF Conference on Computer Vision and Pattern Recognition}}.
  \bibinfo{pages}{17918--17928}.
\newblock


\bibitem[Thoppilan et~al\mbox{.}(2022)]%
        {thoppilan2022lamda}
\bibfield{author}{\bibinfo{person}{Romal Thoppilan}, \bibinfo{person}{Daniel
  De~Freitas}, \bibinfo{person}{Jamie Hall}, \bibinfo{person}{Noam Shazeer},
  \bibinfo{person}{Apoorv Kulshreshtha}, \bibinfo{person}{Heng-Tze Cheng},
  \bibinfo{person}{Alicia Jin}, \bibinfo{person}{Taylor Bos},
  \bibinfo{person}{Leslie Baker}, \bibinfo{person}{Yu Du}, {et~al\mbox{.}}}
  \bibinfo{year}{2022}\natexlab{}.
\newblock \showarticletitle{Lamda: Language models for dialog applications}.
\newblock \bibinfo{journal}{\emph{arXiv preprint arXiv:2201.08239}}
  (\bibinfo{year}{2022}).
\newblock


\bibitem[Tipping and Bishop(1999)]%
        {tipping1999probabilistic}
\bibfield{author}{\bibinfo{person}{Michael~E Tipping} {and}
  \bibinfo{person}{Christopher~M Bishop}.} \bibinfo{year}{1999}\natexlab{}.
\newblock \showarticletitle{Probabilistic principal component analysis}.
\newblock \bibinfo{journal}{\emph{Journal of the Royal Statistical Society
  Series B: Statistical Methodology}} \bibinfo{volume}{61}, \bibinfo{number}{3}
  (\bibinfo{year}{1999}), \bibinfo{pages}{611--622}.
\newblock


\bibitem[Tsimpoukelli et~al\mbox{.}(2021)]%
        {tsimpoukelli2021multimodal}
\bibfield{author}{\bibinfo{person}{Maria Tsimpoukelli},
  \bibinfo{person}{Jacob~L Menick}, \bibinfo{person}{Serkan Cabi},
  \bibinfo{person}{SM Eslami}, \bibinfo{person}{Oriol Vinyals}, {and}
  \bibinfo{person}{Felix Hill}.} \bibinfo{year}{2021}\natexlab{}.
\newblock \showarticletitle{Multimodal few-shot learning with frozen language
  models}.
\newblock \bibinfo{journal}{\emph{Advances in Neural Information Processing
  Systems}}  \bibinfo{volume}{34} (\bibinfo{year}{2021}),
  \bibinfo{pages}{200--212}.
\newblock


\bibitem[{UN Department of Economic and Social Affairs}(2022)]%
        {un2021sustainable}
\bibfield{author}{\bibinfo{person}{{UN Department of Economic and Social
  Affairs}}.} \bibinfo{year}{2022}\natexlab{}.
\newblock \bibinfo{booktitle}{\emph{The Sustainable Development Goals Report
  2021}}.
\newblock \bibinfo{type}{{T}echnical {R}eport}. \bibinfo{institution}{United
  Nations}.
\newblock
\urldef\tempurl%
\url{https://unstats.un.org/sdgs/report/2021/}
\showURL{%
\tempurl}


\bibitem[Uzkent et~al\mbox{.}(2019)]%
        {uzkent2019learning}
\bibfield{author}{\bibinfo{person}{Burak Uzkent}, \bibinfo{person}{Evan
  Sheehan}, \bibinfo{person}{Chenlin Meng}, \bibinfo{person}{Zhongyi Tang},
  \bibinfo{person}{Marshall Burke}, \bibinfo{person}{David Lobell}, {and}
  \bibinfo{person}{Stefano Ermon}.} \bibinfo{year}{2019}\natexlab{}.
\newblock \showarticletitle{Learning to interpret satellite images using
  wikipedia}. In \bibinfo{booktitle}{\emph{Proceedings of the Twenty-Eighth
  International Joint Conference on Artificial Intelligence}}.
\newblock


\bibitem[Wang et~al\mbox{.}(2018a)]%
        {wang2018deep}
\bibfield{author}{\bibinfo{person}{Anna~X Wang}, \bibinfo{person}{Caelin Tran},
  \bibinfo{person}{Nikhil Desai}, \bibinfo{person}{David Lobell}, {and}
  \bibinfo{person}{Stefano Ermon}.} \bibinfo{year}{2018}\natexlab{a}.
\newblock \showarticletitle{Deep transfer learning for crop yield prediction
  with remote sensing data}. In \bibinfo{booktitle}{\emph{Proceedings of the
  1st ACM SIGCAS Conference on Computing and Sustainable Societies}}.
  \bibinfo{pages}{1--5}.
\newblock


\bibitem[Wang et~al\mbox{.}(2018b)]%
        {10.1145/3184558.3186581}
\bibfield{author}{\bibinfo{person}{Wenshan Wang}, \bibinfo{person}{Su Yang},
  \bibinfo{person}{Zhiyuan He}, \bibinfo{person}{Minjie Wang},
  \bibinfo{person}{Jiulong Zhang}, {and} \bibinfo{person}{Weishan Zhang}.}
  \bibinfo{year}{2018}\natexlab{b}.
\newblock \showarticletitle{Urban Perception of Commercial Activeness from
  Satellite Images and Streetscapes}. In \bibinfo{booktitle}{\emph{Companion
  Proceedings of the The Web Conference 2018}} (Lyon, France)
  \emph{(\bibinfo{series}{WWW '18})}. \bibinfo{publisher}{International World
  Wide Web Conferences Steering Committee}, \bibinfo{address}{Republic and
  Canton of Geneva, CHE}, \bibinfo{pages}{647–654}.
\newblock
\showISBNx{9781450356404}
\urldef\tempurl%
\url{https://doi.org/10.1145/3184558.3186581}
\showDOI{\tempurl}


\bibitem[Wang et~al\mbox{.}(2023)]%
        {wang2023would}
\bibfield{author}{\bibinfo{person}{Xinglei Wang}, \bibinfo{person}{Meng Fang},
  \bibinfo{person}{Zichao Zeng}, {and} \bibinfo{person}{Tao Cheng}.}
  \bibinfo{year}{2023}\natexlab{}.
\newblock \showarticletitle{Where would i go next? large language models as
  human mobility predictors}.
\newblock \bibinfo{journal}{\emph{arXiv preprint arXiv:2308.15197}}
  (\bibinfo{year}{2023}).
\newblock


\bibitem[Wang et~al\mbox{.}(2020)]%
        {wang2020urban2vec}
\bibfield{author}{\bibinfo{person}{Zhecheng Wang}, \bibinfo{person}{Haoyuan
  Li}, {and} \bibinfo{person}{Ram Rajagopal}.} \bibinfo{year}{2020}\natexlab{}.
\newblock \showarticletitle{Urban2vec: Incorporating street view imagery and
  pois for multi-modal urban neighborhood embedding}. In
  \bibinfo{booktitle}{\emph{Proceedings of the AAAI Conference on Artificial
  Intelligence}}, Vol.~\bibinfo{volume}{34}. \bibinfo{pages}{1013--1020}.
\newblock


\bibitem[Wei et~al\mbox{.}(2022)]%
        {wei2022emergent}
\bibfield{author}{\bibinfo{person}{Jason Wei}, \bibinfo{person}{Yi Tay},
  \bibinfo{person}{Rishi Bommasani}, \bibinfo{person}{Colin Raffel},
  \bibinfo{person}{Barret Zoph}, \bibinfo{person}{Sebastian Borgeaud},
  \bibinfo{person}{Dani Yogatama}, \bibinfo{person}{Maarten Bosma},
  \bibinfo{person}{Denny Zhou}, \bibinfo{person}{Donald Metzler},
  {et~al\mbox{.}}} \bibinfo{year}{2022}\natexlab{}.
\newblock \showarticletitle{Emergent abilities of large language models}.
\newblock \bibinfo{journal}{\emph{arXiv preprint arXiv:2206.07682}}
  (\bibinfo{year}{2022}).
\newblock


\bibitem[Xi et~al\mbox{.}(2022)]%
        {xi2022beyond}
\bibfield{author}{\bibinfo{person}{Yanxin Xi}, \bibinfo{person}{Tong Li},
  \bibinfo{person}{Huandong Wang}, \bibinfo{person}{Yong Li},
  \bibinfo{person}{Sasu Tarkoma}, {and} \bibinfo{person}{Pan Hui}.}
  \bibinfo{year}{2022}\natexlab{}.
\newblock \showarticletitle{Beyond the first law of geography: Learning
  representations of satellite imagery by leveraging point-of-interests}. In
  \bibinfo{booktitle}{\emph{Proceedings of the ACM Web Conference 2022}}.
  \bibinfo{pages}{3308--3316}.
\newblock


\bibitem[Xu et~al\mbox{.}(2023)]%
        {xu2023urban}
\bibfield{author}{\bibinfo{person}{Fengli Xu}, \bibinfo{person}{Jun Zhang},
  \bibinfo{person}{Chen Gao}, \bibinfo{person}{Jie Feng}, {and}
  \bibinfo{person}{Yong Li}.} \bibinfo{year}{2023}\natexlab{}.
\newblock \showarticletitle{Urban generative intelligence (ugi): A foundational
  platform for agents in embodied city environment}.
\newblock \bibinfo{journal}{\emph{arXiv preprint arXiv:2312.11813}}
  (\bibinfo{year}{2023}).
\newblock


\bibitem[Yang et~al\mbox{.}(2022a)]%
        {yang2022zero}
\bibfield{author}{\bibinfo{person}{Antoine Yang}, \bibinfo{person}{Antoine
  Miech}, \bibinfo{person}{Josef Sivic}, \bibinfo{person}{Ivan Laptev}, {and}
  \bibinfo{person}{Cordelia Schmid}.} \bibinfo{year}{2022}\natexlab{a}.
\newblock \showarticletitle{Zero-shot video question answering via frozen
  bidirectional language models}.
\newblock \bibinfo{journal}{\emph{Advances in Neural Information Processing
  Systems}}  \bibinfo{volume}{35} (\bibinfo{year}{2022}),
  \bibinfo{pages}{124--141}.
\newblock


\bibitem[Yang et~al\mbox{.}(2022b)]%
        {yang2022duare}
\bibfield{author}{\bibinfo{person}{Jianzhong Yang}, \bibinfo{person}{Xiaoqing
  Ye}, \bibinfo{person}{Bin Wu}, \bibinfo{person}{Yanlei Gu},
  \bibinfo{person}{Ziyu Wang}, \bibinfo{person}{Deguo Xia}, {and}
  \bibinfo{person}{Jizhou Huang}.} \bibinfo{year}{2022}\natexlab{b}.
\newblock \showarticletitle{DuARE: Automatic road extraction with aerial images
  and trajectory data at Baidu maps}. In \bibinfo{booktitle}{\emph{Proceedings
  of the 28th ACM SIGKDD Conference on Knowledge Discovery and Data Mining}}.
  \bibinfo{pages}{4321--4331}.
\newblock


\bibitem[Yeh et~al\mbox{.}(2021)]%
        {yeh2021sustainbench}
\bibfield{author}{\bibinfo{person}{Christopher Yeh}, \bibinfo{person}{Chenlin
  Meng}, \bibinfo{person}{Sherrie Wang}, \bibinfo{person}{Anne Driscoll},
  \bibinfo{person}{Erik Rozi}, \bibinfo{person}{Patrick Liu},
  \bibinfo{person}{Jihyeon Lee}, \bibinfo{person}{Marshall Burke},
  \bibinfo{person}{David~B Lobell}, {and} \bibinfo{person}{Stefano Ermon}.}
  \bibinfo{year}{2021}\natexlab{}.
\newblock \showarticletitle{Sustainbench: Benchmarks for monitoring the
  sustainable development goals with machine learning}.
\newblock \bibinfo{journal}{\emph{arXiv preprint arXiv:2111.04724}}
  (\bibinfo{year}{2021}).
\newblock


\bibitem[Yeh et~al\mbox{.}(2020)]%
        {yeh2020using}
\bibfield{author}{\bibinfo{person}{Christopher Yeh}, \bibinfo{person}{Anthony
  Perez}, \bibinfo{person}{Anne Driscoll}, \bibinfo{person}{George Azzari},
  \bibinfo{person}{Zhongyi Tang}, \bibinfo{person}{David Lobell},
  \bibinfo{person}{Stefano Ermon}, {and} \bibinfo{person}{Marshall Burke}.}
  \bibinfo{year}{2020}\natexlab{}.
\newblock \showarticletitle{Using publicly available satellite imagery and deep
  learning to understand economic well-being in Africa}.
\newblock \bibinfo{journal}{\emph{Nature communications}} \bibinfo{volume}{11},
  \bibinfo{number}{1} (\bibinfo{year}{2020}), \bibinfo{pages}{2583}.
\newblock


\bibitem[You et~al\mbox{.}(2017)]%
        {you2017deep}
\bibfield{author}{\bibinfo{person}{Jiaxuan You}, \bibinfo{person}{Xiaocheng
  Li}, \bibinfo{person}{Melvin Low}, \bibinfo{person}{David Lobell}, {and}
  \bibinfo{person}{Stefano Ermon}.} \bibinfo{year}{2017}\natexlab{}.
\newblock \showarticletitle{Deep gaussian process for crop yield prediction
  based on remote sensing data}. In \bibinfo{booktitle}{\emph{Proceedings of
  the AAAI conference on artificial intelligence}}, Vol.~\bibinfo{volume}{31}.
\newblock


\bibitem[Yu et~al\mbox{.}(2022)]%
        {yu2022coca}
\bibfield{author}{\bibinfo{person}{Jiahui Yu}, \bibinfo{person}{Zirui Wang},
  \bibinfo{person}{Vijay Vasudevan}, \bibinfo{person}{Legg Yeung},
  \bibinfo{person}{Mojtaba Seyedhosseini}, {and} \bibinfo{person}{Yonghui Wu}.}
  \bibinfo{year}{2022}\natexlab{}.
\newblock \showarticletitle{CoCa: Contrastive Captioners are Image-Text
  Foundation Models}.
\newblock \bibinfo{journal}{\emph{Transactions on Machine Learning Research}}
  (\bibinfo{year}{2022}).
\newblock
\showISSN{2835-8856}
\urldef\tempurl%
\url{https://openreview.net/forum?id=Ee277P3AYC}
\showURL{%
\tempurl}


\bibitem[Zeng et~al\mbox{.}(2023)]%
        {zeng2023conzic}
\bibfield{author}{\bibinfo{person}{Zequn Zeng}, \bibinfo{person}{Hao Zhang},
  \bibinfo{person}{Ruiying Lu}, \bibinfo{person}{Dongsheng Wang},
  \bibinfo{person}{Bo Chen}, {and} \bibinfo{person}{Zhengjue Wang}.}
  \bibinfo{year}{2023}\natexlab{}.
\newblock \showarticletitle{Conzic: Controllable zero-shot image captioning by
  sampling-based polishing}. In \bibinfo{booktitle}{\emph{Proceedings of the
  IEEE/CVF Conference on Computer Vision and Pattern Recognition}}.
  \bibinfo{pages}{23465--23476}.
\newblock


\bibitem[Zhang et~al\mbox{.}(2021)]%
        {zhang2021vinvl}
\bibfield{author}{\bibinfo{person}{Pengchuan Zhang}, \bibinfo{person}{Xiujun
  Li}, \bibinfo{person}{Xiaowei Hu}, \bibinfo{person}{Jianwei Yang},
  \bibinfo{person}{Lei Zhang}, \bibinfo{person}{Lijuan Wang},
  \bibinfo{person}{Yejin Choi}, {and} \bibinfo{person}{Jianfeng Gao}.}
  \bibinfo{year}{2021}\natexlab{}.
\newblock \showarticletitle{Vinvl: Making visual representations matter in
  vision-language models}.
\newblock \bibinfo{journal}{\emph{arXiv preprint arXiv:2101.00529}}
  \bibinfo{volume}{1}, \bibinfo{number}{6} (\bibinfo{year}{2021}),
  \bibinfo{pages}{8}.
\newblock


\bibitem[Zhang et~al\mbox{.}(2022)]%
        {zhang2022opt}
\bibfield{author}{\bibinfo{person}{Susan Zhang}, \bibinfo{person}{Stephen
  Roller}, \bibinfo{person}{Naman Goyal}, \bibinfo{person}{Mikel Artetxe},
  \bibinfo{person}{Moya Chen}, \bibinfo{person}{Shuohui Chen},
  \bibinfo{person}{Christopher Dewan}, \bibinfo{person}{Mona Diab},
  \bibinfo{person}{Xian Li}, \bibinfo{person}{Xi~Victoria Lin},
  {et~al\mbox{.}}} \bibinfo{year}{2022}\natexlab{}.
\newblock \showarticletitle{Opt: Open pre-trained transformer language models}.
\newblock \bibinfo{journal}{\emph{arXiv preprint arXiv:2205.01068}}
  (\bibinfo{year}{2022}).
\newblock


\bibitem[Zhang et~al\mbox{.}(2024)]%
        {zhang2024towards}
\bibfield{author}{\bibinfo{person}{Weijia Zhang}, \bibinfo{person}{Jindong
  Han}, \bibinfo{person}{Zhao Xu}, \bibinfo{person}{Hang Ni},
  \bibinfo{person}{Hao Liu}, {and} \bibinfo{person}{Hui Xiong}.}
  \bibinfo{year}{2024}\natexlab{}.
\newblock \showarticletitle{Towards Urban General Intelligence: A Review and
  Outlook of Urban Foundation Models}.
\newblock \bibinfo{journal}{\emph{arXiv preprint arXiv:2402.01749}}
  (\bibinfo{year}{2024}).
\newblock


\bibitem[Zhao et~al\mbox{.}(2023)]%
        {zhao2023survey}
\bibfield{author}{\bibinfo{person}{Wayne~Xin Zhao}, \bibinfo{person}{Kun Zhou},
  \bibinfo{person}{Junyi Li}, \bibinfo{person}{Tianyi Tang},
  \bibinfo{person}{Xiaolei Wang}, \bibinfo{person}{Yupeng Hou},
  \bibinfo{person}{Yingqian Min}, \bibinfo{person}{Beichen Zhang},
  \bibinfo{person}{Junjie Zhang}, \bibinfo{person}{Zican Dong},
  {et~al\mbox{.}}} \bibinfo{year}{2023}\natexlab{}.
\newblock \showarticletitle{A survey of large language models}.
\newblock \bibinfo{journal}{\emph{arXiv preprint arXiv:2303.18223}}
  (\bibinfo{year}{2023}).
\newblock


\bibitem[Zhou et~al\mbox{.}(2023)]%
        {zhou2023one}
\bibfield{author}{\bibinfo{person}{Tian Zhou}, \bibinfo{person}{Peisong Niu},
  \bibinfo{person}{Xue Wang}, \bibinfo{person}{Liang Sun}, {and}
  \bibinfo{person}{Rong Jin}.} \bibinfo{year}{2023}\natexlab{}.
\newblock \showarticletitle{One Fits All: Power General Time Series Analysis by
  Pretrained LM}.
\newblock \bibinfo{journal}{\emph{Advances in Neural Information Processing
  Systems}} (\bibinfo{year}{2023}).
\newblock


\end{thebibliography}

\appendix

\section{More Related Work}\label{app:related}
\subsection{Vision-Language Pre-Training (VLP)}

VLP aims for effective vision-language alignment with frozen unimodal models from the vision and natural language communities. CLIP \cite{radford2021learning} is proposed as a vision foundation model based on image-text contrastive learning rationale. In certain approaches, the image encoder is frozen to extract visual features, as exemplified by the frozen object detector \cite{zhang2021vinvl} or the pre-trained image encoder for CLIP used in LiT \cite{zhang2022opt}. In contrast, some methods freeze the language model to leverage the knowledge from LLMs for vision-to-language generation tasks. To align visual features with the fixed text space, Frozen \cite{tsimpoukelli2021multimodal} fine-tunes an image encoder whose outputs are fed as soft prompts for LLMs. Flamingo \cite{dai2022enabling} pre-trains a cross-attention layer added into the LLM to inject visual features. Moreover, BLIP-2 \cite{li2023blip} takes full advantage of both frozen image encoders and frozen LLMs for various vision-language tasks. Leveraging the VLP scheme, we align generated text with satellite images to produce interpretable representations for urban regions.

\subsection{Urban Foundation Model (UFM)}

UFMs represent a novel family of models pre-trained on extensive urban data sources, encompassing multi-granularity data and multi-modality \cite{zhang2024towards,tang2024synergizing}. Language-based UFMs fall within two categories: pre-training models on geo-text information \cite{huang2022ernie,ding2023mgeo} and adaptation of existing LLMs to urban scenarios \cite{manvi2023geollm}. Similarly, vision-based UFMs can follow the same category - pre-training approach \cite{wang2020urban2vec,liu2023knowledge} and adaptation approach \cite{sultan2023geosam}. Furthermore, UFMs tend to cover time series \cite{jintime2023}, trajectory \cite{wang2023would}, and multimodal domains \cite{xu2023urban,hao2024urbanvlp}.

\section{Image-to-Text Foundation Models}\label{app: image2text}

We provide a brief introduction of the Image-to-Text foundation models that we used for text generation:
\begin{itemize}[left=0pt]
    \item \textbf{BLIP}. A VLP framework that leverages noisy web data for bootstrapping captions; it involves a captioner generating synthetic captions and a filter to eliminate noisy ones.
    \item \textbf{Emu}. A multimodal foundation model trained with a unified objective, either classifying the next text token or regressing the next visual embedding in the multimodal sequence.
    \item \textbf{ImageBind-LLM}. A multimodal instruction model that unifies various modalities such as images and video into a single framework by aligning ImageBind's visual encoder with an LLM using a learnable bind network.
    \item \textbf{PandaGPT}. A unified approach that can handle multimodal inputs, allowing composition of their semantics by combining multimodal encoders from ImageBind and LLMs from Vicuna.
    \item \textbf{OpenFlamingo}. An open-source multimodal framework that is capable of handling diverse visual language tasks through autoregressive vision-language modeling.
    \item \textbf{mPLUG}. A VLP model with an efficient vision-language architecture, equipped with innovative cross-modal skip-connections.
    \item \textbf{LLaVA}. An instruction tuning-based model that utilizes multimodal language-image data derived from GPT4.
\end{itemize}

\section{Details of Text Refinement}\label{app: refinement}

Text refinement is of critical importance to the language-image pertaining framework of our proposed UrbanCLIP. In particular, we defined two key types of low-quality text:
\begin{itemize}[left=0pt]
    \item \textbf{Unfactual information}. For example, a satellite image lacking any water bodies contains one description: "The image features a large body of water, possibly a river or a lake, running through the city". Such incongruent text can disrupt the alignment between visual and textual modalities.
    \item \textbf{Vague expression}. As an illustration, consider a satellite image accompanied by the description: "The image offers a comprehensive view of the city's layout and infrastructure". Such text may not contribute beneficially to the valid fusion of LLM-inherent knowledge into an image encoder.
\end{itemize}

Due to the well-known "hallucination" issue, LLMs are susceptible to generating unfactual text descriptions. To this end, we devise a two-stage heuristic process for text refinement in this paper, encompassing text cleaning and counterfactual verification.
\begin{itemize}[left=0pt]
    \item \textbf{Text cleaning}. In the first stage, we adopt a rule-based approach, utilizing pre-configured regular filters to eliminate redundant and irrelevant textual information. Popular text processing tools from the NLTK package \footnote{\url{https://github.com/nltk/nltk}} are employed to remove noise.
    \item \textbf{Counterfactual verification}. In the second stage, following the acquisition of high-quality image-text pairs, we enlist the expertise of many Master students with dual backgrounds in GIS and CS for factual verification. To enhance the accuracy of this process, an auxiliary dual-scoring mechanism is deployed to filter out anomalous descriptive information.
\end{itemize}

Additionally, we have explored the application of the text discriminator model trained from BLIP \cite{li2022blip} for automated scoring and filtering. However, in comparison to our two-stage heuristic approach, the automated filtering process encounters challenges related to performance instability and the production of low-quality filter results. Therefore, we consider this motivation as part of our future work, as detailed in Section \ref{conclusion}, and encourage researchers to delve into the exploration of automatic, high-quality text refinement processes.

\section{Complexity Analysis}\label{app: complexity}
We use the following notations: \( m_1 \) represents the number of visual tokens of ViT, \( d \) is the dimension of the representation, \( L \) denotes the number of layers in the transformer (assuming uniformity across ViT, textual transformer, and multimodal transformer), and \( m_2 \) stands for the sequence length of textual tokens. 

\begin{figure*}[!b]
  \centering
  \vspace*{0.4cm}
  \includegraphics[width=0.9\textwidth]{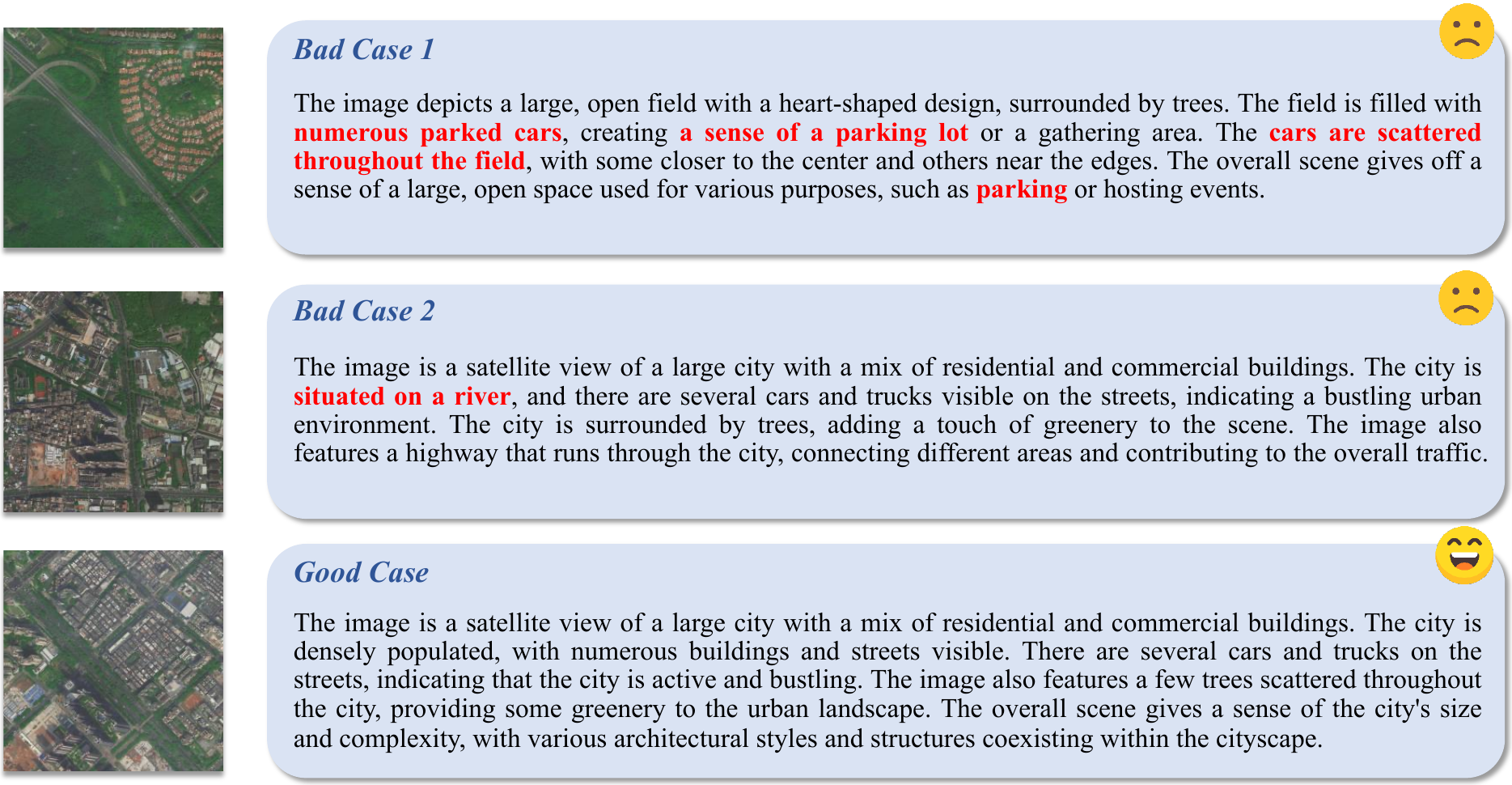}
  \caption{Bad and good case examples of text description generated by LLaMA-Adapter V2.}
  \label{fig: text_example}
\end{figure*}

For the visual encoder, the complexity of ViT is \(\mathbf{O}(L (m_1^2 d + m_1 d^2))\) and that of attentional pooling is \(\mathbf{O}(m_1^2 d)\). The textual encoder has an embedding lookup complexity of \(\mathbf{O}(m_2 d)\) and transformers with \(\mathbf{O}(L (m_2^2 d + m_2 d^2))\). The multimodal interaction involves cross attention with a complexity of \(\mathbf{O}(L m_1 m_2 d)\). The final complexity, when summing up, is \(\mathbf{O}(L (m_1^2 + m_2^2) d + m_1^2 d + L m_1 m_2 d)\), which for large values of \( m_1n \) and \( m_2 \) approximates to \(\mathbf{O}(L (m_1^2 d + m_2^2 d))\). Besides, LLM pre-training is excluded from UrbanCLIP backbone training, and text generation and refinement remain at the preprocessing phase, thus indicating the feasibility of UrbanCLIP in practice.

\section{LLM Limitation Analysis}\label{app: bad case}
As illustrated in Figure \ref{fig: text_example}, we incorporated additional instances where LLM may not generate texts effectively. This inclusion aims to provide a more comprehensive understanding of the capabilities and limitations of LLM. Based on our observations, we have identified two common bad cases that illustrate these challenges.
\begin{itemize}[left=0pt]
    \item \textbf{Bad case 1}. If the road network is complex, particularly with intersections within residential areas, there is a potential for LLM to erroneously assume the presence of numerous parked cars along the road.
    \item \textbf{Bad case 2}. If there is a highway crossing through the residential area, LLM may misinterpret it and consider it as a river within the urban region.
\end{itemize}
The reasons for the aforementioned issues can be summarized as follows. On one hand, some information from satellite images is \textbf{inherently challenging for the human eye} to discern. On the other hand, there could be the "\textbf{hallucination}" issue of LLM, which is restricted by the capacity bottleneck of LLaMA-based models.


\end{document}